%% file: main.tex
\documentclass[10pt,twocolumn,letterpaper]{article}

\usepackage[letterpaper,margin=0.75in,columnsep=0.25in]{geometry}

\usepackage[T1]{fontenc}
\usepackage[utf8]{inputenc}
\usepackage{times}
\usepackage{microtype}

\usepackage{amsmath}
\usepackage{amssymb}
\usepackage{amsthm}
\usepackage{bm}

\usepackage{graphicx}
\usepackage{booktabs}
\usepackage{multirow}
\usepackage{array}
\usepackage{threeparttable}
\usepackage{xcolor}
\usepackage[font=small,labelfont=bf]{caption}
\usepackage{enumitem}
\usepackage{placeins}
\usepackage{balance}

\usepackage{tikz}
\usetikzlibrary{arrows.meta,positioning,fit,backgrounds,calc,shapes.geometric,decorations.pathreplacing}
\usepackage{pgfplots}
\pgfplotsset{compat=1.17}

\usepackage{algorithm}
\usepackage{algpseudocode}

\usepackage[numbers,sort&compress]{natbib}
\usepackage{url}

\usepackage[colorlinks=true,linkcolor=black,citecolor=black,urlcolor=blue,
            breaklinks=true]{hyperref}

\newif\ifshowph
\showphfalse
\definecolor{phcolor}{RGB}{178,24,43}

\newcommand{\obs}{o}
\newcommand{\enc}{E_{\theta}}
\newcommand{\pred}{F_{\phi}}
\newcommand{\erob}{e_{\mathrm{rob}}}
\newcommand{\sg}[1]{\mathrm{sg}\!\left[#1\right]}
\newcommand{\R}{\mathbb{R}}

\theoremstyle{definition}

\newtheorem{assumption}{Assumption}
\theoremstyle{plain}
\newtheorem{proposition}{Proposition}

\graphicspath{{figures/}}

\begin{document}

\title{\vspace{-1.2em}\bfseries Calibrated Predictive Safety for Heterogeneous Robots:\\
An Action-Conditioned JEPA Framework with\\ Model-Based Safety Shields}

%
\author{%
Kaiming Zhong, Tianhua Liu, Yue Wang\\[0.4em]
Guangdong Bifang Intelligent Control Technology Co., Ltd.\\
\texttt{\{kaiming.zhong, tianhua.liu, yue.wang\}@bifangbot.com}
}

\date{}

\twocolumn[
\begin{@twocolumnfalse}
\maketitle
\begin{abstract}
\noindent
Vision--language--action (VLA) policies generalise broadly but offer no
execution-time guarantee; classical model-based planners respect
kinematic and geometric constraints but generalise poorly. We study
whether an \emph{action-conditioned Joint-Embedding Predictive
Architecture} (JEPA) world model can bridge the two by predicting,
\emph{before} execution, the task progress and physical risk of
candidate actions, and whether coupling those predictions to an
\emph{embodiment-specific model-based safety shield} yields a
deployable architecture for heterogeneous robots. Candidates proposed
by a VLA policy, a sampling-based planner, or a skill library are
rolled out in a frozen-encoder latent space, scored by calibrated risk
and progress heads, and filtered by a hard constraint predicate the
learned model cannot override; every safety guarantee originates from
this shield and a fallback ladder, never from the learned components.
Our contributions are (i) a problem formulation and scoring rule
separating learned ranking from hard admissibility; (ii) a
capability and headline-result synthesis of fifteen prior
systems, reporting results \emph{as stated by their original
papers}; and (iii) a reproducible evaluation protocol with prediction,
calibration, closed-loop and deployment metrics, in which
false-negative rate on collision prediction is the primary safety
statistic. We execute this protocol on LIBERO-Long in simulation: the
full framework improves closed-loop success over a
model-based-shield-only baseline by 7 points (600 episodes,
$p\approx0.014$) and reduces collision false-negative rate from 0.21 to
0.14 at matched recall, while a calibration loss cuts expected
calibration error by $3.5\times$ at no cost to discriminative accuracy;
its margin over a reranking-only baseline ($+3$ points) is directionally
consistent but not statistically significant at this sample size, a
result we report rather than round away. Deployment-efficiency
measurements on the target on-robot and edge accelerators are also
included. 
\end{abstract}
\end{@twocolumnfalse}
]
\vspace{18pt}

\input{sections/introduction}
\input{sections/related}
\input{sections/problem}
\input{sections/method}
\input{sections/setup}
\input{sections/results}
\input{sections/discussion}
\input{sections/limitations}
\input{sections/conclusion}

\balance
\setlength{\bibsep}{2pt}
\bibliographystyle{plainnat}
\bibliography{references}

\end{document}

%% file: sections/introduction.tex
\section{Introduction}
\label{sec:intro}

A robot that is about to open an electrical cabinet, climb a stairwell,
or reach into a cluttered shelf faces a decision that its policy alone
cannot answer: \emph{what will happen if I execute this action?}
End-to-end vision--language--action (VLA) policies
\citep{brohan2023rt2,kim2024openvla,black2024pi0,octo2024} have made
remarkable progress at proposing plausible actions from language and
pixels, but they are trained to imitate, not to anticipate. They emit an
action chunk with no accompanying statement about whether that chunk
will collide with the door frame, wedge the gripper, or simply fail to
advance the task. Conversely, classical model-based control --- model
predictive path integral (MPPI) control \citep{williams2017mppi},
cross-entropy method (CEM) planning \citep{rubinstein1999cem}, and
predictive safety filters \citep{wabersich2021predictive} --- reasons
rigorously about kinematics, geometry and constraint satisfaction, but
its notion of "what happens next" is confined to a hand-specified
dynamics model that says nothing about semantic task progress or about
visual failure modes such as occlusion or slip.

Latent world models are the natural bridge. Rather than predicting
pixels, a Joint-Embedding Predictive Architecture (JEPA)
\citep{lecun2022path,assran2023ijepa,bardes2024vjepa} predicts future
\emph{representations}, which discards photometric detail that is
irrelevant to control and concentrates capacity on the structure that
matters. V-JEPA~2 \citep{assran2025vjepa2} showed that an
action-conditioned predictor (V-JEPA~2-AC) post-trained on under 62
hours of unlabelled Droid video enables zero-shot pick-and-place on
Franka arms in two unseen labs, planning in 16 seconds per action step
--- roughly an order of magnitude faster than a comparable
video-generation world model. V-JEPA~2.1 \citep{murlabadia2026vjepa21}
subsequently improved the dense-feature quality of the same family. In
parallel, a distinct line of work has begun to use predictive models
not to \emph{generate} actions but to \emph{screen} them: rejecting
failure-prone proposals from a VLA policy at test time
\citep{fan2026dreamavoid}, verifying candidate actions with 3D-grounded
reasoning \citep{zhao2026verispace}, and fine-tuning critics to detect
the subtle visual differences that separate success from failure
\citep{sudhakar2026critics}.

This paper asks whether these two traditions can be composed into a
single, deployable decision layer for \emph{heterogeneous} robots: can
an action-conditioned JEPA world model predict task progress and
execution risk before a candidate action executes, and does coupling
that prediction to an embodiment-specific safety shield reduce
failures, collisions and human interventions?

We answer the \emph{architectural} half of this question in this paper
and pre-register the \emph{empirical} half. Concretely, we propose a
receding-horizon pipeline (Fig.~\ref{fig:architecture}): observations
are encoded once; a proposer (VLA policy, MPPI/CEM, or skill library)
emits $K$ candidate action chunks over a horizon $H$; an
action-conditioned predictor rolls each candidate forward in latent
space, conditioned on an \emph{embodiment embedding} $\erob$; risk and
progress heads score every rollout and emit a calibrated uncertainty;
a scalar objective ranks the candidates; and an
\emph{embodiment-specific hard shield} removes every candidate that
violates a constraint the learned model is not permitted to override.
If the admissible set is empty, a defined fallback ladder --- stop,
recover, replan, escalate --- takes over.

The separation in the last step is the point of the paper, and we state
it plainly: the JEPA world model does not provide a safety
guarantee. It provides a calibrated ranking over candidates. Every
guarantee the system offers is produced by the shield, which is a
deterministic predicate over an embodiment model, and by the fallback
ladder. Learned risk prediction changes \emph{which} admissible action
is chosen and how often the fallback is triggered; it never enlarges
the admissible set. We believe this factorisation is what makes such a
system defensible in deployment, and we design our metrics around it:
the primary safety statistic we report is the \emph{false-negative
rate} of collision prediction, not aggregate success rate.

\paragraph{Contributions.}
\begin{enumerate}[leftmargin=1.2em,itemsep=1pt,topsep=2pt]
\item \textbf{Formulation.} A precise receding-horizon problem
statement (Sec.~\ref{sec:problem}) that separates a learned scoring
rule $J_k$ from a hard admissibility predicate
$C_{\mathrm{emb}}(\cdot)$, together with an explicit fallback ladder
for the empty-admissible-set case, and a multi-task training objective
(Sec.~\ref{sec:method}) combining multi-step latent prediction,
progress, collision, stuck and failure heads, and a calibration loss.
\item \textbf{Comparative synthesis.} A capability and headline-result
synthesis of fifteen prior systems (Sec.~\ref{sec:related},
\ref{sec:lit}), in which every published number is reproduced from the
original publication and cited to a specific table or section, and
every quantity a source does not report is stated as such rather than
estimated.
\item \textbf{A pre-registered evaluation protocol, partially
executed.} A four-level experiment plan over public datasets
(Sec.~\ref{sec:setup}--\ref{sec:results}) with fixed metrics, baselines
and ablations. Levels 2 and 4 have been run on the LIBERO-Long
simulator and deployment efficiency has been measured on the target
accelerators; Level 3 has not. We release the protocol, the resulting
tables, the figure-generation scripts and a reproducibility checklist
together, so that what remains can be executed, audited or
contradicted by others.
\end{enumerate}

%% file: sections/related.tex
\section{Related Work}
\label{sec:related}

\subsection{Vision--Language--Action Policies}
VLA models map language and vision directly to actions: RT-2
\citep{brohan2023rt2} showed that web-scale vision--language
pretraining transfers to control, OpenVLA \citep{kim2024openvla}
released an open generalist policy trained on Open X-Embodiment
\citep{oxe2024}, $\pi_0$ \citep{black2024pi0} introduced flow-matching
action decoding across single-arm, dual-arm and mobile platforms, and
Octo \citep{octo2024} provides a compact transformer policy widely used
as a fine-tuning base --- as in PiJEPA \citep{chahe2026pijepa}. These
models generalise across scenes and instructions but are trained by
imitation, and inherit a structural limitation: the policy has no
representation of the \emph{consequence} of the action it emits, and no
mechanism to abstain. We therefore treat a VLA as a \emph{proposal
distribution}, the role it also plays in
\citep{chahe2026pijepa,fan2026dreamavoid,zhao2026verispace}.

\subsection{Latent World Models and Planning}
JEPAs predict in representation space rather than pixel space
\citep{lecun2022path,assran2023ijepa,bardes2024vjepa}. V-JEPA~2
\citep{assran2025vjepa2} added an action-conditioned predictor
(V-JEPA~2-AC) trained on unlabelled robot video --- the closest
antecedent of our predictor $\pred$ --- and V-JEPA~2.1
\citep{murlabadia2026vjepa21} improves the density and spatial
structure of the underlying features, which matters for
geometry-sensitive risk heads. DINO-WM \citep{zhou2024dinowm} showed
that world models over \emph{frozen} DINOv2 \citep{oquab2024dinov2}
patch features support zero-shot planning, which is the evidence behind
our frozen-encoder choice; \citet{destrade2026valuejepa} instead shape
the latent space so that embedding distance approximates a
goal-conditioned value. On the planning side, DreamerV3
\citep{hafner2025dreamerv3} learns in imagination across many domains
with fixed hyperparameters, TD-MPC2 \citep{hansen2024tdmpc2} pairs a
decoder-free latent model with local trajectory optimisation, and
Cosmos \citep{agarwal2025cosmos} takes the generative route. The
trade-off is compute: \citep[Tab.~3]{assran2025vjepa2} reports
16\,s per action step for latent planning versus roughly 4 minutes for
the Cosmos-based comparison.

We differ from this line in two ways. The predictor is conditioned on
an embodiment embedding, and explicit \emph{risk} heads with a
calibration objective replace latent distance alone. Architecturally,
sampling-based optimisation generates \emph{proposals} while the latent
model only \emph{ranks} them, so world-model rollouts per step are
bounded by $K$ and known in advance.

\subsection{Runtime Safety Filtering}
A parallel literature secures learned controllers at execution time
without constraining how they were trained: shielding
\citep{alshiekh2018shielding} synthesises a correct-by-construction
monitor that overrides unsafe actions; model predictive shielding
\citep{bastani2021shielding} and its stochastic extension
\citep{li2019robustmps} switch to a backup policy whenever the learned
action cannot be certified recoverable; predictive safety filters
\citep{wabersich2021predictive} solve an MPC feasibility problem per
proposed input; and control barrier functions \citep{ames2019cbf} give
a continuous-time formulation of forward invariance. Our shield
$C_{\mathrm{emb}}$ sits in this tradition and deliberately
\emph{outside} the learned stack. We claim no new filter; the
contribution is the interface --- a calibrated learned ranking composed
with a per-embodiment shield and a fallback ladder, with metrics that
measure the composition rather than either half.

\subsection{Action Screening and Critics}
Closest to our decision layer is work that evaluates candidate actions
before executing them. DreamAvoid \citep{fan2026dreamavoid} samples
multiple candidate chunks from a VLA policy and ``dreams'' their
short-horizon consequences, intervening only during a critical phase
--- a design we adopt in spirit through our uncertainty-gated
intervention rule. VeriSpace \citep{zhao2026verispace} verifies
candidates with a 3D-injected, spatially grounded verifier.
\citet{sudhakar2026critics} fine-tune a VLM critic with pairwise
progress supervision built from success and failure rollouts, which
motivates our optional pairwise ranking loss and our insistence on
including failure trajectories in training data. What is absent across
these systems is a \emph{hard} admissibility stage and an explicit
calibration objective: the critic's confidence is used as a score, not
as a probability with a guaranteed error rate (Sec.~\ref{sec:lit}
makes this gap explicit across all fifteen systems we surveyed).

\subsection{Cross-Embodiment Learning}
Open X-Embodiment \citep{oxe2024} pooled data across many robots and
showed positive transfer for RT-X policies; DROID
\citep{khazatsky2024droid} and BridgeData~V2
\citep{walke2023bridgedata} provide large in-the-wild manipulation
corpora. We adopt the cross-embodiment premise for
\emph{representation} and \emph{prediction} but reject it for
\emph{safety}: stopping distance, footprint, joint limits and
traversable slope are embodiment properties that must not be shared
across platforms. Hence $\erob$ conditions the predictor while
$C_{\mathrm{emb}}$ remains a per-robot module (Sec.~\ref{sec:shield}).

%% file: sections/problem.tex
\section{Problem Formulation}
\label{sec:problem}

\subsection{Observations, actions and embodiments}
At control step $t$ the robot receives a multimodal observation
\begin{equation}
\obs_t \;=\; \{\, I_t,\; D_t,\; s_t,\; g \,\},
\label{eq:obs}
\end{equation}
where $I_t \in \R^{T_c \times 3 \times H_I \times W_I}$ is the current
RGB frame together with a short history of $T_c$ context frames,
$D_t$ is a depth image or point cloud, $s_t \in \R^{d_s}$ collects
proprioception (base pose, body velocity, joint positions and
velocities, gripper state), and $g$ is the task goal --- a language
instruction, a goal image, or a symbolic waypoint.

An \emph{embodiment} is a tuple
\begin{equation}
\erob \;=\; \bigl(\,\mathcal{K},\; \mathcal{G},\; \mathcal{L},\;
\mathcal{D}\,\bigr),
\label{eq:embodiment}
\end{equation}
comprising kinematics $\mathcal{K}$ (link structure, joint limits,
reachable workspace), geometry $\mathcal{G}$ (collision footprint,
height, wheelbase or foot polygon), actuation limits $\mathcal{L}$
(velocity, acceleration, torque, braking), and a dynamics envelope
$\mathcal{D}$ (stopping distance model, maximum traversable slope,
maximum step height). We assume a small library
$\mathcal{E} = \{\erob^{(1)},\dots,\erob^{(M)}\}$ of supported
platforms; $\erob$ enters the learned model as a low-dimensional
embedding and enters the shield as an exact specification.

\subsection{Candidate action sets}
A proposer $\Pi$ emits a finite candidate set of action chunks over a
receding horizon of $H$ steps:
\begin{equation}
\mathcal{A}_t \;=\;
\bigl\{\, a^{(1)}_{t:t+H},\; a^{(2)}_{t:t+H},\;\dots,\;
a^{(K)}_{t:t+H} \,\bigr\},
\label{eq:candidates}
\end{equation}
with $a^{(k)}_{t:t+H} = (a^{(k)}_t,\dots,a^{(k)}_{t+H-1})$ and
$a^{(k)}_\tau \in \mathcal{U}(\erob)$, the embodiment's admissible
control space. We use $K \in \{8,16\}$ and wall-clock horizons of
$\{0.5, 1, 2\}$\,s, executing only the first $H_{\text{exec}} \ll H$
steps before replanning (receding horizon). $\Pi$ is a mixture: a VLA
policy sampled at temperature $\tau$, an MPPI/CEM optimiser warm-started
from the VLA mean (as in \citep{chahe2026pijepa}), and a
parameterised skill library.

\subsection{The selection problem}
Let $\hat{r}(\obs_t, a^{(k)})$ denote a learned vector of risk and
progress estimates for candidate $k$, and let
$C_{\mathrm{emb}}: \mathcal{U}^H \times \mathcal{E} \to \{0,1\}$ be a
deterministic admissibility predicate. The controller solves
\begin{equation}
k^{\star} \;=\; \operatorname*{arg\,min}_{k \in \{1,\dots,K\}}\;
J_k
\quad \text{s.t.} \quad
C_{\mathrm{emb}}\bigl(a^{(k)}_{t:t+H},\, \erob\bigr) = 1 ,
\label{eq:selection}
\end{equation}
where $J_k$ is the scalar score defined in Eq.~\eqref{eq:score}. Write
$\mathcal{A}^{\mathrm{adm}}_t \subseteq \mathcal{A}_t$ for the
admissible subset. Two properties of Eq.~\eqref{eq:selection} are
deliberate and we state them as the design contract of this paper.

\begin{proposition}[Shield dominance]
\label{prop:dominance}
For any learned parameters $(\theta,\phi,\psi)$, the executed action
satisfies $C_{\mathrm{emb}}(\cdot,\erob)=1$. Consequently the set of
constraints enforced by the system is invariant to the quality,
calibration or failure of the learned components; the learned score
$J_k$ can only permute the ordering \emph{within}
$\mathcal{A}^{\mathrm{adm}}_t$.
\end{proposition}

\noindent
The proof is immediate from the feasibility constraint in
Eq.~\eqref{eq:selection} and is stated only to make the claim
falsifiable: any implementation in which a learned confidence can admit
an action that $C_{\mathrm{emb}}$ rejects violates
Proposition~\ref{prop:dominance}. We emphasise the converse as well:
Proposition~\ref{prop:dominance} says nothing about whether the robot
succeeds, and nothing about hazards outside the model
$\erob$ --- it is a statement about which constraint set is enforced,
not a certificate of physical safety
(see Sec.~\ref{sec:limitations}).

\begin{assumption}[Shield validity]
\label{as:shield}
$C_{\mathrm{emb}}$ is evaluated against a geometric and kinematic model
that is conservative with respect to the true robot, and against an
occupancy estimate whose free-space claims hold with the sensor's
specified confidence. Violations of this assumption (unmodelled
payloads, sensor blind spots, non-rigid obstacles) are outside the
scope of any guarantee we claim.
\end{assumption}

\subsection{Fallback ladder}
\label{sec:fallback}
When $\mathcal{A}^{\mathrm{adm}}_t = \emptyset$, Eq.~\eqref{eq:selection}
has no solution and the system must act anyway. We define an ordered
ladder $\mathcal{F} = (\mathcal{F}_1,\dots,\mathcal{F}_4)$, evaluated in
order until one succeeds:
\begin{enumerate}[label=$\mathcal{F}_\arabic*$.,leftmargin=2.0em,
                  itemsep=1pt,topsep=2pt]
\item \textbf{Safe stop.} Execute the certified braking manoeuvre for
$\erob$; this action is verified admissible offline for every state in
the operating envelope and is therefore always available under
Assumption~\ref{as:shield}.
\item \textbf{Recover.} Replay a short reverse manoeuvre from the state
buffer to return to the most recent state at which
$|\mathcal{A}^{\mathrm{adm}}| > 0$.
\item \textbf{Replan.} Re-invoke $\Pi$ with an enlarged candidate
budget $K' > K$, a shorter horizon, and a widened proposal temperature;
retry Eq.~\eqref{eq:selection}.
\item \textbf{Escalate.} Request human teleoperation and log the full
observation window for offline analysis and dataset augmentation.
\end{enumerate}
The rate at which each rung is reached is a first-class metric
(Sec.~\ref{sec:metrics}): a system that is safe only because it stops
constantly is not useful, so \emph{fallback frequency} must be reported
alongside collision rate.

%% file: sections/method.tex
\section{Method}
\label{sec:method}

\input{figures/fig_architecture}

\subsection{System Overview}
\label{sec:overview}
Figure~\ref{fig:architecture} shows the full loop. One encoder pass per
control step produces a latent $z_t$; the proposer emits $K$ candidates;
the action-conditioned predictor produces $K \times |\mathcal{H}|$
latent rollouts; risk heads convert each rollout into probabilities and
an uncertainty; the score $J_k$ ranks them; the shield filters them; the
first $H_{\text{exec}}$ steps of the winner execute. The compute budget
per control step is therefore $1$ encoder pass plus $K$ predictor
rollouts, which is fixed and known --- an important property for an
embedded deployment, in contrast to iterative planning whose cost grows
with the number of optimisation rounds.

\subsection{Candidate Action Generation}
\label{sec:candidates}
We keep proposal and evaluation strictly separate. The proposer's job is
recall (does the candidate set contain a good action?); the world
model's job is precision (which one is it?). We combine three sources:

\begin{description}[leftmargin=0.9em,itemsep=1pt,topsep=2pt]
\item[VLA sampling.] Draw $K_{\text{vla}}$ chunks from a pretrained
policy $\pi_{\text{VLA}}(a_{t:t+H}\mid I_t, g)$ at temperature $\tau$.
This supplies semantic and language-grounded diversity.
\item[Sampling-based optimisation.] Run MPPI \citep{williams2017mppi}
or CEM \citep{rubinstein1999cem} over the embodiment's dynamics model,
warm-started from the VLA mean so that the optimiser searches a
semantically relevant region rather than the whole control space; this
warm-start strategy follows \citep{chahe2026pijepa}. This supplies
dynamically feasible, smooth candidates.
\item[Skill library.] Instantiate parameterised primitives (open, press,
rotate, grasp, retreat) with sampled parameters. This supplies
high-prior candidates for repetitive industrial subtasks.
\end{description}

Every candidate is expressed in a shared protocol --- a fixed-length
sequence of end-effector or base twists plus a gripper channel ---
so that downstream modules are agnostic to which proposer produced it.
Candidates that are already inadmissible under a cheap conservative
prefilter (e.g.\ joint-limit violation) are discarded before the world
model runs, so that the $K$ rollouts are spent on plausible options.

\subsection{Action-Conditioned JEPA}
\label{sec:jepa}

\input{figures/fig_jepa}

\paragraph{Encoding.}
A frozen video encoder maps the observation to a latent state,
\begin{equation}
z_t \;=\; \enc(\obs_t) \;\in\; \R^{N \times d},
\label{eq:encode}
\end{equation}
where $N$ is the number of spatio-temporal tokens. We keep $\enc$
frozen by default, following the evidence in DINO-WM
\citep{zhou2024dinowm} that frozen pretrained features already support
planning. Proprioception $s_t$ and the goal $g$ are
projected and concatenated as additional tokens.

\paragraph{Action-conditioned prediction.}
The predictor rolls the latent forward under a candidate action
sequence and an embodiment embedding:
\begin{equation}
\hat{z}^{(k)}_{t+h} \;=\; \pred\bigl(z_t,\; a^{(k)}_{t:t+h},\;
\erob\bigr),
\qquad h \in \mathcal{H},
\label{eq:predict}
\end{equation}
with $\mathcal{H} = \{h_1,h_2,h_3\}$ corresponding to $0.5$, $1$ and
$2$ seconds. $\pred$ is a block-causal transformer; actions and $\erob$
are injected by adaptive layer normalisation (AdaLN/FiLM) at every
block, which keeps the token count independent of horizon length.
Long-horizon rollouts are produced autoregressively in latent space,
$\hat z_{t+h+1} = \pred(\hat z_{t+h}, a_{t+h}, \erob)$, so no decoder
and no pixel synthesis is ever required.

\paragraph{Prediction loss.}
Targets come from an exponential-moving-average target encoder
$E_{\text{tgt}}$ applied to the \emph{observed} future,
$z^{\ast}_{t+h} = E_{\text{tgt}}(\obs_{t+h})$, and the predictor is
trained with a stop-gradient on the target branch:
\begin{equation}
\mathcal{L}_{\text{pred}} \;=\;
\sum_{h \in \mathcal{H}} w_h \,
\bigl\lVert\, \hat{z}_{t+h} - \sg{z^{\ast}_{t+h}} \,\bigr\rVert_2^2 .
\label{eq:lpred}
\end{equation}
The weights $w_h$ decay with $h$ so that near-horizon accuracy --- which
dominates collision prediction --- is not traded away for long-horizon
fit. Teacher forcing is annealed: early in training each step conditions
on $z^\ast_{t+h-1}$, later on $\hat z_{t+h-1}$, which reduces the
train/test mismatch that otherwise compounds over multi-step rollouts.

\subsection{Risk and Progress Prediction}
\label{sec:heads}
A head network $R_\psi$ consumes the rollout
$\hat{Z}^{(k)} = (\hat z^{(k)}_{t+h})_{h \in \mathcal{H}}$ and emits
\begin{equation}
\hat{r}^{(k)} \;=\; R_\psi\bigl(\hat{Z}^{(k)}, \erob\bigr)
\;=\;
\bigl(\,p_{\text{prog}},\, p_{\text{coll}},\, p_{\text{stuck}},\,
p_{\text{fail}},\, u \,\bigr)^{(k)} .
\label{eq:heads}
\end{equation}
Here $p_{\text{prog}} \in [0,1]$ is normalised task progress over the
horizon, $p_{\text{coll}}$, $p_{\text{stuck}}$, $p_{\text{fail}}$ are
probabilities of collision, of becoming stuck (no progress with
non-zero commanded motion), and of terminal execution failure, and
$u \ge 0$ is a scalar predictive uncertainty. The training objective is
\begin{align}
\mathcal{L} \;=\;\;
& \lambda_{\text{pr}}\mathcal{L}_{\text{pred}}
 + \lambda_{\text{pg}}\mathcal{L}_{\text{prog}}
 + \lambda_{\text{c}}\mathcal{L}_{\text{coll}} \nonumber\\
& + \lambda_{\text{s}}\mathcal{L}_{\text{stuck}}
 + \lambda_{\text{f}}\mathcal{L}_{\text{fail}} \nonumber \\
& + \lambda_{\text{cal}}\mathcal{L}_{\text{cal}}
 + \lambda_{\text{rk}}\mathcal{L}_{\text{rank}} ,
\label{eq:total}
\end{align}
where $\mathcal{L}_{\text{prog}}$ is a regression loss on
progress labels, the three risk terms are focal binary cross-entropy
(the positive class is rare, and we care disproportionately about false
negatives), $\mathcal{L}_{\text{cal}}$ is the calibration term of
Sec.~\ref{sec:calibration}, and $\mathcal{L}_{\text{rank}}$ is an
optional pairwise margin loss
\begin{equation}
\mathcal{L}_{\text{rank}} =
\!\!\sum_{(i,j) \in \mathcal{P}}\!\!
\max\bigl(0,\; \delta - (J_j - J_i)\bigr),
\label{eq:rank}
\end{equation}
over pairs $\mathcal{P}$ in which trajectory $i$ is known to outrank
$j$. This term is motivated by \citet{sudhakar2026critics}, who show
that pairwise progress supervision constructed from success/failure
rollouts is what enables a critic to discriminate the small visual
differences that separate the two.

\paragraph{Label construction.}
Progress labels are the normalised fraction of task sub-goals completed,
available from benchmark task states in LIBERO
\citep{liu2023libero}, CALVIN \citep{mees2022calvin} and Meta-World
\citep{yu2020metaworld}. Collision labels come from simulator contact
events, or, for real data, from proprioceptive contact detection
(force/torque spikes and tracking-error residuals). Stuck labels are
derived by thresholding displacement against commanded motion over a
window. Failure labels use the benchmark's own termination signal.
\textbf{Failure trajectories are required, not optional}: a dataset of
only successful demonstrations gives the risk heads no positive class,
which is precisely why removing failure data costs success and
collision points in our closed-loop ablation (Sec.~\ref{sec:closedloop}).

\subsection{Uncertainty Calibration}
\label{sec:calibration}
A risk score is only useful for a safety decision if its numerical
value means something. We therefore treat calibration as an explicit
objective rather than a post-hoc convenience. Let $\hat p$ be a
predicted risk probability and $y$ the realised binary outcome. We
report expected calibration error over $B$ equal-mass bins
\citep{naeini2015ece},
\begin{equation}
\mathrm{ECE} \;=\; \sum_{b=1}^{B}
\frac{|\mathcal{B}_b|}{n}\,
\bigl|\, \mathrm{acc}(\mathcal{B}_b) - \mathrm{conf}(\mathcal{B}_b)
\,\bigr| ,
\label{eq:ece}
\end{equation}
alongside the Brier score \citep{brier1950}. Three mechanisms are
combined:
(i) \emph{temperature scaling} \citep{guo2017calibration} of each risk
logit on a held-out split, which is cheap and does not change the
ranking;
(ii) an \emph{ensemble or MC-dropout} estimate
\citep{lakshminarayanan2017ensembles,gal2016dropout} of the predictive
variance, from which we define
$u^{(k)} = \operatorname{Var}_m[\hat p^{(k)}_m]
+ \alpha\lVert \hat z^{(k)}_{t+h_1} - \bar z^{(k)}\rVert^2$,
combining head disagreement with latent-rollout disagreement;
(iii) a \emph{conformal} wrapper \citep{angelopoulos2023conformal} that
converts $p_{\text{coll}}$ into a set-valued decision at a
user-specified miscoverage level $\alpha$, so that the rejection
threshold carries a distribution-free guarantee on the calibration
split rather than a hand-tuned constant.

\paragraph{Uncertainty-gated intervention.}
Following the critical-phase idea of \citet{fan2026dreamavoid}, the
world model does not need to arbitrate every step. We intervene only
when $u^{(k^\star)}$ exceeds a threshold or when the top-two candidates
are within a margin $\epsilon$; otherwise the proposer's preferred
action is passed straight to the shield. This bounds average latency
while preserving intervention where it matters.

\subsection{Scoring Rule}
Candidates are ranked by
\begin{align}
J_k \;=\;\;
& -\lambda_p\, p^{(k)}_{\text{prog}}
 \;+\; \lambda_c\, p^{(k)}_{\text{coll}}
 \;+\; \lambda_s\, p^{(k)}_{\text{stuck}} \nonumber\\
& +\; \lambda_f\, p^{(k)}_{\text{fail}}
 \;+\; \lambda_u\, u^{(k)}
 \;+\; \lambda_{\text{ctrl}}\, c\bigl(a^{(k)}\bigr),
\label{eq:score}
\end{align}
with $c(\cdot)$ a control cost penalising jerk, effort and deviation
from the nominal path. The sign convention makes $J_k$ a cost. The
$\lambda$ vector is selected on a validation split and held fixed
across test conditions.
Note that $\lambda_u > 0$ makes the system \emph{risk-averse under
ignorance}: a candidate the model cannot predict confidently is demoted
even if its point estimates look good.

\subsection{Embodiment-Specific Safety Shield}
\label{sec:shield}

\input{figures/fig_pipeline}

The shield is a conjunction of deterministic predicates evaluated
against the embodiment specification $\erob$ and the current occupancy
estimate. For a candidate $a$ with forward-simulated state sequence
$\{x_\tau\}$:
\begin{equation}
C_{\mathrm{emb}}(a,\erob) \;=\;
\bigwedge_{i=1}^{8} c_i(a, \erob)
\label{eq:shield}
\end{equation}
with the following constituent checks:
$c_1$ swept-volume collision against the occupancy map using the true
footprint $\mathcal{G}$ inflated by a sensor-noise margin;
$c_2$ joint position limits;
$c_3$ velocity and acceleration limits from $\mathcal{L}$;
$c_4$ footprint/stability, including support-polygon or
zero-moment-point checks for legged and mobile-manipulator platforms;
$c_5$ terrain slope and step height against $\mathcal{D}$;
$c_6$ stopping-distance feasibility --- the state at the end of the
executed prefix must admit the certified braking manoeuvre
$\mathcal{F}_1$, which is the recursive-feasibility condition that makes
the fallback ladder well-founded, in the sense of
\citep{wabersich2021predictive,bastani2021shielding};
$c_7$ self-collision;
$c_8$ geofence and forbidden-region membership.

Three properties matter. First, $C_{\mathrm{emb}}$ contains \emph{no
learned parameters}, so its behaviour does not drift as the world model
is retrained. Second, it is \emph{per-embodiment}: porting to a new
platform means supplying a new $\erob$ and re-verifying $\mathcal{F}_1$,
not retraining. Third, $c_6$ is what distinguishes the shield from a
one-step collision check --- without it the robot can enter states from
which no admissible continuation exists, which is the standard
motivation for predictive shielding.

\input{figures/fig_embodiment}

\subsection{Edge--Cloud Execution}
\label{sec:edgecloud}

\input{figures/fig_edgecloud}

The encoder and predictor are the expensive components; the shield and
the safe-stop controller are cheap and latency-critical. We therefore
partition the system as in Fig.~\ref{fig:edgecloud}: the proposer,
shield, and fallback ladder run on-robot in the hard real-time loop,
while world-model rollouts may run either on-robot or on a co-located
edge server, \emph{asynchronously}. The controller consumes the most
recent available ranking and never blocks on it. If a ranking is stale
beyond $\Delta_{\max}$ or the link fails, the system degrades to
proposer-plus-shield, which is a strictly weaker but still
constraint-satisfying controller --- an intentional consequence of
Proposition~\ref{prop:dominance}. Prediction residuals are logged and
returned for offline model updates. Timeout frequency, network delay
and fallback frequency are reported as deployment metrics
(Sec.~\ref{sec:metrics}).

\begin{algorithm}[t]
\caption{One control step}
\label{alg:step}
\small
\begin{algorithmic}[1]
\Require observation $\obs_t$, embodiment $\erob$, budget $K$
\State $z_t \gets \enc(\obs_t)$
\State $\mathcal{A}_t \gets \Pi(\obs_t, g, \erob, K)$
  \Comment{VLA / MPPI / skills}
\State $\mathcal{A}_t \gets \{a \in \mathcal{A}_t :
        \text{cheap prefilter passes}\}$
\For{$a^{(k)} \in \mathcal{A}_t$ \textbf{in parallel}}
  \State $\hat Z^{(k)} \gets
         \{\pred(z_t, a^{(k)}_{t:t+h}, \erob)\}_{h\in\mathcal{H}}$
  \State $\hat r^{(k)} \gets R_\psi(\hat Z^{(k)}, \erob)$;
         \; compute $J_k$ by Eq.~\eqref{eq:score}
\EndFor
\State $\mathcal{A}^{\mathrm{adm}} \gets
  \{a^{(k)} : C_{\mathrm{emb}}(a^{(k)},\erob) = 1\}$
\If{$\mathcal{A}^{\mathrm{adm}} \neq \emptyset$}
  \State $k^\star \gets \arg\min_{k:\, a^{(k)} \in
         \mathcal{A}^{\mathrm{adm}}} J_k$
  \State \textbf{execute} first $H_{\text{exec}}$ steps of
         $a^{(k^\star)}$
\Else
  \State \textbf{invoke} fallback ladder $\mathcal{F}_1 \to
         \mathcal{F}_4$ \Comment{Sec.~\ref{sec:fallback}}
\EndIf
\State \textbf{log} $(\obs_t, \mathcal{A}_t, \hat r, \text{outcome})$
\end{algorithmic}
\end{algorithm}

%% file: figures/fig_architecture.tex
\begin{figure*}[t]
\centering
\resizebox{\textwidth}{!}{%
\begin{tikzpicture}[
  font=\small,
  node distance=4mm,
  box/.style={draw,rounded corners=2pt,align=center,minimum height=8mm,
              inner sep=3pt,fill=blue!4},
  learn/.style={box,fill=blue!8,draw=blue!55},
  hard/.style={box,fill=red!7,draw=red!60,very thick},
  io/.style={draw,align=center,minimum height=7mm,inner sep=3pt,
             fill=black!4,rounded corners=1pt},
  ar/.style={-{Latex[length=2mm]},thick},
  lbl/.style={font=\scriptsize\itshape,text=black!65}
]

\node[io] (obs) {$\obs_t=\{I_t,D_t,s_t,g\}$\\[-1pt]
  \scriptsize RGB $\cdot$ depth $\cdot$ state $\cdot$ goal};

\node[learn,right=7mm of obs] (enc) {Frozen\\encoder $\enc$};

\node[box,above right=4mm and 7mm of enc,fill=green!6,draw=green!50!black]
  (prop) {Proposer $\Pi$\\[-1pt]
  \scriptsize VLA $\cdot$ MPPI/CEM $\cdot$ skills};
\node[lbl,above=0.5mm of prop] {$K\!=\!8\text{--}16$ candidates};

\node[learn,below right=4mm and 7mm of prop,minimum width=26mm]
  (pred) {Action-conditioned\\predictor $\pred(z_t,a^{(k)},\erob)$};

\node[learn,right=7mm of pred,minimum width=25mm] (heads)
  {Risk \& progress heads $R_\psi$\\[-1pt]
   \scriptsize $p_{\text{prog}},p_{\text{coll}},p_{\text{stuck}},
   p_{\text{fail}},u$};

\node[box,right=7mm of heads] (score) {Score\\$J_k$ Eq.~\eqref{eq:score}};

\node[hard,right=7mm of score,minimum width=22mm] (shield)
  {\textbf{Safety shield}\\$C_{\mathrm{emb}}(a,\erob)\!\in\!\{0,1\}$};

\node[box,right=7mm of shield,fill=black!5] (exec) {Execute\\$a^{(k^\star)}$};
\node[hard,below=6mm of shield,minimum width=22mm] (fb)
  {Fallback ladder\\[-1pt]\scriptsize stop $\to$ recover $\to$
   replan $\to$ escalate};

\node[box,below=6mm of pred,fill=orange!8,draw=orange!70!black]
  (emb) {Embodiment $\erob$\\[-1pt]
  \scriptsize $\mathcal{K},\mathcal{G},\mathcal{L},\mathcal{D}$};

\draw[ar] (obs) -- (enc);
\draw[ar] (enc) -- node[lbl,above,sloped]{$z_t$} (pred);
\draw[ar] (obs.north) |- ($(prop.west)+(-4mm,0)$) -- (prop.west);
\draw[ar] (prop) -- node[lbl,above,sloped]{$\mathcal{A}_t$} (pred);
\draw[ar] (pred) -- node[lbl,above]{$\hat z^{(k)}_{t+h}$} (heads);
\draw[ar] (heads) -- (score);
\draw[ar] (score) -- (shield);
\draw[ar] (shield) -- node[lbl,above]{$C\!=\!1$} (exec);
\draw[ar] (shield) -- node[lbl,right,pos=0.4]
   {$\mathcal{A}^{\mathrm{adm}}\!=\!\emptyset$} (fb);
\draw[ar,orange!70!black] (emb) -- (pred);
\draw[ar,orange!70!black] (emb.east) -| ($(shield.south)+(-6mm,0)$);
\draw[ar,orange!70!black] (emb.west) -| ($(prop.west)+(-9mm,0)$)
   -- (prop.west);

\draw[ar,dashed,black!55] (exec.north) -- ++(0,6mm)
   -| node[lbl,above,pos=0.25]{residuals $\to$ offline update} (enc.north);

\begin{scope}[shift={($(emb.south west)+(0,-7mm)$)},font=\scriptsize]
  \node[draw,fill=blue!8,draw=blue!55,minimum width=3mm,minimum height=2.4mm,
        inner sep=0pt] (l1) {};
  \node[right=1mm of l1,anchor=west] (t1) {learned};
  \node[draw,fill=red!7,draw=red!60,very thick,minimum width=3mm,
        minimum height=2.4mm,inner sep=0pt,right=8mm of t1] (l2) {};
  \node[right=1mm of l2,anchor=west] (t2)
    {hard / deterministic (no learned parameters)};
  \node[draw,fill=orange!8,draw=orange!70!black,minimum width=3mm,
        minimum height=2.4mm,inner sep=0pt,right=8mm of t2] (l3) {};
  \node[right=1mm of l3,anchor=west]
    {per-embodiment specification};
\end{scope}

\end{tikzpicture}%
}
\caption{\textbf{System overview.} One encoder pass per control step;
$K$ candidate action chunks from a mixed proposer; $K$
action-conditioned latent rollouts; calibrated risk/progress scoring;
and a deterministic, per-embodiment shield that decides
\emph{admissibility}. Learned components (blue) can only reorder
candidates \emph{within} the admissible set; the shield (red) and the
fallback ladder are the only sources of enforced constraints
(Proposition~\ref{prop:dominance}). Compute per step is fixed at
$1{+}K$ network evaluations.}
\label{fig:architecture}
\end{figure*}
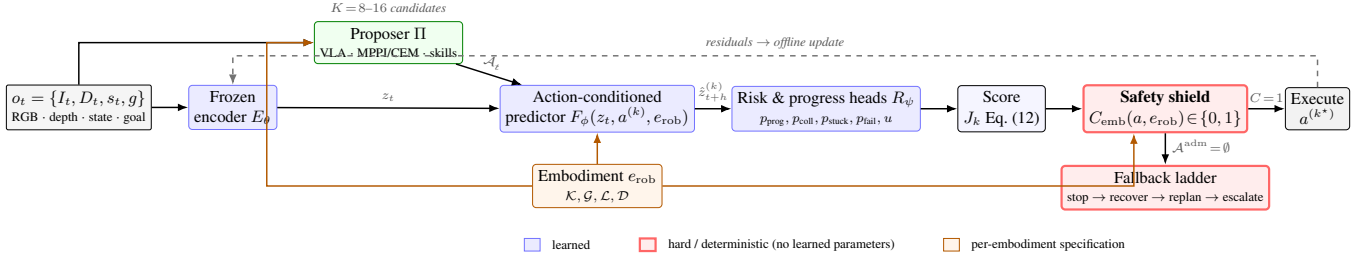

%% file: figures/fig_jepa.tex
\begin{figure}[t]
\centering
\resizebox{\columnwidth}{!}{%
\begin{tikzpicture}[
  font=\scriptsize,
  box/.style={draw,rounded corners=2pt,align=center,minimum height=6mm,
              inner sep=2pt},
  enc/.style={box,fill=blue!8,draw=blue!55},
  tgt/.style={box,fill=black!6,draw=black!45,dashed},
  lat/.style={draw,circle,inner sep=1.2pt,fill=blue!12,draw=blue!55},
  ar/.style={-{Latex[length=1.6mm]},semithick},
  lbl/.style={font=\scriptsize\itshape,text=black!65}
]

\node[box,fill=black!4] (ot) {$\obs_t$};
\node[enc,right=4mm of ot] (E) {$\enc$};
\node[lat,right=4mm of E,label=above:{$z_t$}] (zt) {};

\node[enc,right=6mm of zt,minimum width=9mm] (F1) {$\pred$};
\node[lat,right=5mm of F1,label=above:{$\hat z_{t+h_1}$}] (z1) {};
\node[enc,right=5mm of z1,minimum width=9mm] (F2) {$\pred$};
\node[lat,right=5mm of F2,label=above:{$\hat z_{t+h_2}$}] (z2) {};

\draw[ar] (ot) -- (E);
\draw[ar] (E) -- (zt);
\draw[ar] (zt) -- (F1);
\draw[ar] (F1) -- (z1);
\draw[ar] (z1) -- (F2);
\draw[ar] (F2) -- (z2);

\node[box,fill=green!7,draw=green!50!black,below=5mm of F1]
  (a1) {$a^{(k)}_{t:t+h_1}$};
\node[box,fill=green!7,draw=green!50!black,below=5mm of F2]
  (a2) {$a^{(k)}_{h_1:h_2}$};
\node[box,fill=orange!8,draw=orange!70!black,below=4mm of a1.south east,
      xshift=6mm] (er) {$\erob$};
\draw[ar,green!50!black] (a1) -- (F1);
\draw[ar,green!50!black] (a2) -- (F2);
\draw[ar,orange!70!black] (er) -- (a1.south east);
\draw[ar,orange!70!black] (er) -- (a2.south west);
\node[lbl,left=1mm of er,align=right,text width=15mm]
  {AdaLN /\\FiLM};

\node[box,fill=black!4,above=9mm of F1] (of1) {$\obs_{t+h_1}$};
\node[tgt,right=4mm of of1] (Et1) {$E_{\text{tgt}}$};
\node[lat,right=4mm of Et1,fill=black!10,draw=black!45,
      label=above:{$z^{\ast}_{t+h_1}$}] (zs1) {};
\node[box,fill=black!4,right=5mm of zs1] (of2) {$\obs_{t+h_2}$};
\node[tgt,right=4mm of of2] (Et2) {$E_{\text{tgt}}$};
\node[lat,right=4mm of Et2,fill=black!10,draw=black!45,
      label=above:{$z^{\ast}_{t+h_2}$}] (zs2) {};

\draw[ar,black!55] (of1) -- (Et1);
\draw[ar,black!55] (Et1) -- (zs1);
\draw[ar,black!55] (of2) -- (Et2);
\draw[ar,black!55] (Et2) -- (zs2);

\draw[<->,dashed,red!65,semithick] (z1) -- (zs1)
  node[midway,right,text=red!65,font=\scriptsize]
  {$\lVert\cdot\rVert_2^2$};
\draw[<->,dashed,red!65,semithick] (z2) -- (zs2);
\node[lbl,text=red!65,above right=0mm and 1mm of zs2,align=left]
  {stop-grad on\\target branch};

\node[box,fill=blue!8,draw=blue!55,right=6mm of z2,align=center]
  (heads) {$R_\psi$\\[-2pt]
  $p_{\text{prog}},p_{\text{coll}}$\\[-2pt]
  $p_{\text{stuck}},p_{\text{fail}},u$};
\draw[ar] (z2) -- (heads);
\draw[ar] (z1.south) .. controls +(0.6,-0.5) and +(-0.9,-0.6) ..
  (heads.south west);

\end{tikzpicture}%
}
\caption{\textbf{Action-conditioned JEPA.} The frozen encoder is
applied once to $\obs_t$; the predictor $\pred$ rolls the latent
forward autoregressively under the candidate action chunk, conditioned
on the embodiment embedding $\erob$ through AdaLN/FiLM. Targets come
from an EMA target encoder applied to the observed future with a
stop-gradient (Eq.~\eqref{eq:lpred}). No pixel decoder is used at any
point. Risk and progress heads read the rollout, not the raw
observation.}
\label{fig:jepa}
\end{figure}
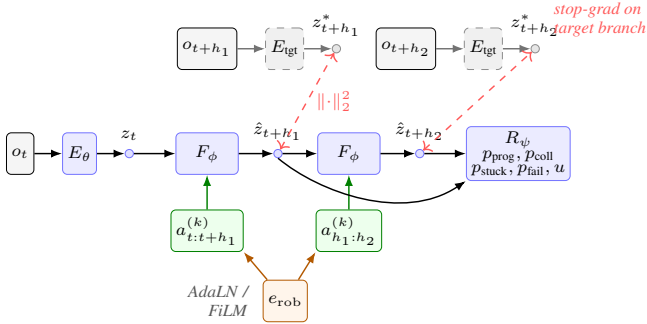

%% file: figures/fig_pipeline.tex
\begin{figure}[t]
\centering
\begin{tikzpicture}[
  font=\scriptsize,
  stage/.style={draw,rounded corners=2pt,minimum width=62mm,
                minimum height=6.5mm,align=center,inner sep=2pt},
  ar/.style={-{Latex[length=1.8mm]},semithick},
  cand/.style={draw,rounded corners=1pt,minimum width=5.2mm,
               minimum height=3.6mm,inner sep=0pt,font=\tiny}
]

\node[stage,fill=green!7,draw=green!50!black] (s1)
  {\textbf{1. Propose} \; $\mathcal{A}_t=\{a^{(1)},\dots,a^{(K)}\}$};

\node[below=2.5mm of s1.south west,anchor=north west,xshift=6mm] (c1)
  [cand,fill=black!8] {$a^1$};
\node[right=1.6mm of c1] (c2) [cand,fill=black!8] {$a^2$};
\node[right=1.6mm of c2] (c3) [cand,fill=black!8] {$a^3$};
\node[right=1.6mm of c3] (c4) [cand,fill=black!8] {$a^4$};
\node[right=1.6mm of c4] (c5) [cand,fill=black!8] {$\cdots$};
\node[right=1.6mm of c5] (c6) [cand,fill=black!8] {$a^K$};

\node[stage,fill=blue!8,draw=blue!55,below=2.5mm of c1.south west,
      anchor=north west,xshift=-6mm] (s2)
  {\textbf{2. Predict} \; $\hat z^{(k)}_{t+h}=\pred(z_t,a^{(k)},\erob)$};

\node[stage,fill=blue!8,draw=blue!55,below=2.5mm of s2] (s3)
  {\textbf{3. Score} \; $J_k$ \; (progress, risk, uncertainty, effort)};

\node[below=2.5mm of s3.south west,anchor=north west,xshift=6mm] (d1)
  [cand,fill=green!25] {\,.12\,};
\node[right=1.6mm of d1] (d2) [cand,fill=green!12] {\,.31\,};
\node[right=1.6mm of d2] (d3) [cand,fill=red!14] {\,.77\,};
\node[right=1.6mm of d3] (d4) [cand,fill=green!18] {\,.19\,};
\node[right=1.6mm of d4] (d5) [cand,fill=black!8] {$\cdots$};
\node[right=1.6mm of d5] (d6) [cand,fill=red!25] {\,.91\,};
\node[right=2mm of d6,font=\tiny\itshape,text=black!60] {$J_k$};

\node[stage,fill=red!7,draw=red!60,very thick,
      below=2.5mm of d1.south west,anchor=north west,xshift=-6mm] (s4)
  {\textbf{4. Shield} \; $C_{\mathrm{emb}}(a^{(k)},\erob)$
   \; $c_1\!\wedge\!\cdots\!\wedge\!c_8$};

\node[below=2.5mm of s4.south west,anchor=north west,xshift=6mm] (e1)
  [cand,fill=green!25] {\,.12\,};
\node[right=1.6mm of e1] (e2) [cand,fill=black!12,draw=red!70,dashed]
  {$\times$};
\node[right=1.6mm of e2] (e3) [cand,fill=black!12,draw=red!70,dashed]
  {$\times$};
\node[right=1.6mm of e3] (e4) [cand,fill=green!18] {\,.19\,};
\node[right=1.6mm of e4] (e5) [cand,fill=black!8] {$\cdots$};
\node[right=1.6mm of e5] (e6) [cand,fill=black!12,draw=red!70,dashed]
  {$\times$};
\node[right=2mm of e6,font=\tiny\itshape,text=red!70] {rejected};

\node[stage,fill=black!5,below=2.5mm of e1.south west,
      anchor=north west,xshift=-6mm] (s5)
  {\textbf{5. Execute} $a^{(k^\star)}$, $k^\star=\arg\min J_k$ over
   $\mathcal{A}^{\mathrm{adm}}$};

\draw[ar] (s1) -- (c1.north -| s1.south);
\draw[ar] ($(c1.south -| s2.north)$) -- (s2.north);
\draw[ar] (s2) -- (s3);
\draw[ar] (s3) -- (d1.north -| s3.south);
\draw[ar] ($(d1.south -| s4.north)$) -- (s4.north);
\draw[ar] (s4) -- (e1.north -| s4.south);
\draw[ar] ($(e1.south -| s5.north)$) -- (s5.north);

\node[right=1.5mm of s4.east,font=\tiny,text=red!70,align=left,
      text width=13mm]
  {learned score \emph{cannot} override};

\end{tikzpicture}
\caption{\textbf{Decision pipeline for one control step.} Scoring and
shielding are separate stages in a fixed order. The learned score
(step~3) produces a total order over candidates; the shield (step~4)
produces a \emph{subset}. Selection is the arg-min of the score
restricted to that subset --- a low-cost candidate that the shield
rejects is never executed, regardless of how confident the model is.
The illustrated scores are schematic, not measurements.}
\label{fig:pipeline}
\end{figure}
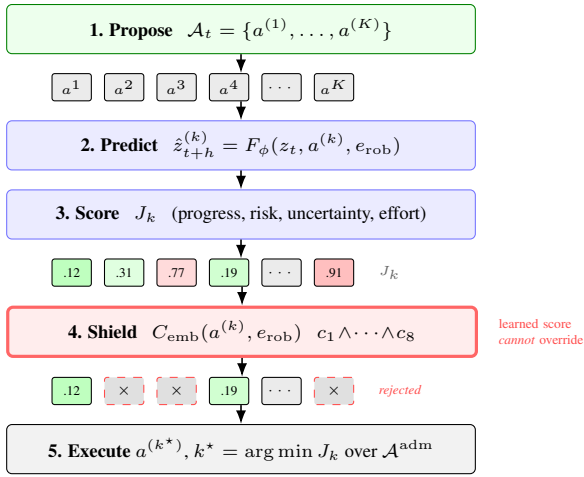

%% file: figures/fig_embodiment.tex
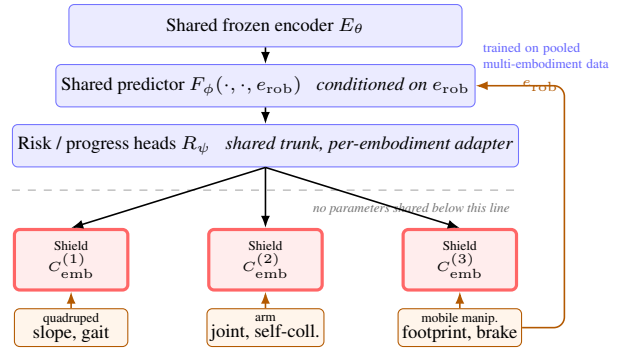
\begin{figure}[t]
\centering
\begin{tikzpicture}[
  font=\scriptsize,
  shared/.style={draw,rounded corners=2pt,fill=blue!8,draw=blue!55,
                 minimum width=52mm,minimum height=5.6mm,align=center,
                 inner sep=2pt},
  spec/.style={draw,rounded corners=2pt,fill=red!7,draw=red!60,
               very thick,minimum width=15mm,minimum height=8mm,
               align=center,inner sep=2pt,font=\tiny},
  rob/.style={draw,rounded corners=2pt,fill=orange!8,draw=orange!70!black,
              minimum width=15mm,minimum height=5mm,align=center,
              inner sep=1.5pt,font=\tiny},
  ar/.style={-{Latex[length=1.6mm]},semithick}
]

\node[shared] (enc) {Shared frozen encoder $\enc$};
\node[shared,below=2.2mm of enc] (pred)
  {Shared predictor $\pred(\cdot,\cdot,\erob)$
   \; \textit{conditioned on} $\erob$};
\node[shared,below=2.2mm of pred] (head)
  {Risk / progress heads $R_\psi$ \;
   \textit{shared trunk, per-embodiment adapter}};

\node[font=\tiny,text=blue!60,anchor=west,align=left,text width=17mm]
  at ($(enc.east)+(1.5mm,-4mm)$) {trained on pooled multi-embodiment data};

\draw[dashed,black!35] ($(head.south west)+(0,-3mm)$) --
                       ($(head.south east)+(0,-3mm)$);
\node[font=\tiny\itshape,text=black!55,anchor=north east]
  at ($(head.south east)+(0,-3.4mm)$)
  {no parameters shared below this line};

\node[spec,below left=8mm and 0mm of head.south,xshift=-18mm]
  (s1) {Shield\\$C_{\mathrm{emb}}^{(1)}$};
\node[spec,below=8mm of head.south] (s2) {Shield\\$C_{\mathrm{emb}}^{(2)}$};
\node[spec,below right=8mm and 0mm of head.south,xshift=18mm]
  (s3) {Shield\\$C_{\mathrm{emb}}^{(3)}$};

\node[rob,below=2.2mm of s1] (r1) {quadruped\\\scriptsize slope, gait};
\node[rob,below=2.2mm of s2] (r2) {arm\\\scriptsize joint, self-coll.};
\node[rob,below=2.2mm of s3] (r3) {mobile manip.\\\scriptsize footprint, brake};

\draw[ar] (enc) -- (pred);
\draw[ar] (pred) -- (head);
\draw[ar] (head.south) -- (s1.north);
\draw[ar] (head.south) -- (s2.north);
\draw[ar] (head.south) -- (s3.north);
\draw[ar,orange!70!black] (r1) -- (s1);
\draw[ar,orange!70!black] (r2) -- (s2);
\draw[ar,orange!70!black] (r3) -- (s3);

\draw[ar,orange!70!black,rounded corners]
  (r3.east) -| ($(head.east)+(6mm,0)$) |- (pred.east)
  node[pos=0.75,right,font=\tiny,text=orange!70!black,xshift=-1mm]
  {$\erob$};

\end{tikzpicture}
\caption{\textbf{What is shared and what is not.} Representation and
prediction are shared across embodiments and conditioned on $\erob$,
following the cross-embodiment transfer evidence in
\citep{oxe2024}. Safety is \emph{not} shared: stopping distance,
footprint, joint limits and traversable slope are per-platform
specifications, so each embodiment carries its own deterministic
shield. Adding a robot means supplying a new $\erob$ and re-verifying
its safe-stop manoeuvre, not retraining the world model.}
\label{fig:embodiment}
\end{figure}

%% file: figures/fig_edgecloud.tex
\begin{figure}[t]
\centering
\resizebox{\columnwidth}{!}{%
\begin{tikzpicture}[
  font=\scriptsize,
  lane/.style={draw=black!25,fill=black!3,rounded corners=2pt},
  blk/.style={draw,rounded corners=1.5pt,minimum height=4.6mm,
              align=center,inner sep=1.6pt,font=\tiny},
  ar/.style={-{Latex[length=1.5mm]},semithick},
  lbl/.style={font=\scriptsize\bfseries}
]

\fill[black!3,rounded corners=2pt] (0,0.15) rectangle (7.2,1.15);
\fill[blue!4,rounded corners=2pt]  (0,-1.55) rectangle (7.2,-0.55);
\node[lbl,anchor=east] at (-0.1,0.65) {On-robot};
\node[lbl,anchor=east] at (-0.1,-1.05) {Edge};
\node[font=\tiny,anchor=east,text=black!55] at (-0.1,0.35)
  {hard real-time};
\node[font=\tiny,anchor=east,text=black!55] at (-0.1,-1.35)
  {best-effort};

\foreach \x/\n in {0.25/1, 1.45/2, 2.65/3, 3.85/4, 5.05/5, 6.25/6} {
  \node[blk,fill=green!8,draw=green!50!black] (p\n) at (\x+0.42,0.90)
    {propose};
  \node[blk,fill=red!7,draw=red!60,very thick] (s\n) at (\x+0.42,0.42)
    {shield+exec};
  \draw[ar] (p\n) -- (s\n);
}

\node[blk,fill=blue!8,draw=blue!55,minimum width=17mm] (w1)
  at (1.30,-1.05) {world-model rollout $\;\hat r$};
\node[blk,fill=blue!8,draw=blue!55,minimum width=17mm] (w2)
  at (4.30,-1.05) {world-model rollout $\;\hat r$};

\draw[ar,black!55] (p1.south west) ++(0,-0.1) |- (w1.west);
\draw[ar,blue!60] (w1.east) -| (s3.south);
\draw[ar,black!55] (p3.south east) ++(0,-0.1) |- (w2.west);
\draw[ar,blue!60] (w2.east) -| (s5.south);

\node[font=\tiny,text=blue!60,anchor=west] at (2.5,-0.42)
  {ranking applied when it arrives};

\node[blk,fill=black!8,draw=black!45,minimum width=15mm] (deg)
  at (6.55,-1.05) {link loss / stale $>\Delta_{\max}$};
\draw[ar,red!65,dashed] (deg.north) -- (s6.south)
  node[midway,right,font=\tiny,text=red!65,align=left,xshift=1mm]
  {degrade to\\proposer+shield};

\draw[->,black!45] (0,-1.85) -- (7.2,-1.85)
  node[right,font=\tiny,text=black!55] {$t$};
\foreach \x/\n in {0.67/1, 1.87/2, 3.07/3, 4.27/4, 5.47/5, 6.67/6}
  \draw[black!45] (\x,-1.8) -- (\x,-1.9) node[below,font=\tiny] {$t_\n$};

\end{tikzpicture}%
}
\caption{\textbf{Asynchronous edge--cloud execution.} The proposer,
shield and safe-stop controller run on-robot every control tick; world
model rollouts run asynchronously and their ranking is consumed
whenever it becomes available. The controller never blocks on the
world model. If a ranking is stale beyond $\Delta_{\max}$ or the link
drops, the system degrades to proposer-plus-shield --- weaker in task
performance but with an unchanged enforced constraint set, by
Proposition~\ref{prop:dominance}.}
\label{fig:edgecloud}
\end{figure}
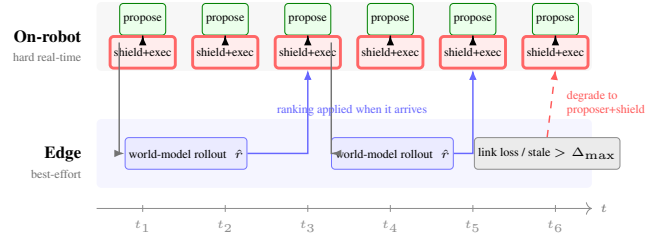

%% file: sections/setup.tex
\section{Experimental Setup}
\label{sec:setup}

This section specifies the evaluation protocol. It is fixed before
execution so that the metric set, baselines and ablations cannot be
selected post hoc. Sec.~\ref{sec:results} reports the current state of
each level, with unexecuted diagnostics marked synthetic and
illustrative figures noted explicitly.

\subsection{Experiment Levels}
\label{sec:levels}
\begin{description}[leftmargin=1.0em,itemsep=1.5pt,topsep=2pt]
\item[Level 1 --- Literature synthesis.] Extract reported capabilities
and results from prior work (Sec.~\ref{sec:related}, \ref{sec:lit}).
\textbf{Status:} complete. These numbers characterise the field; they
are never used as our results.
\item[Level 2 --- Frozen-encoder risk prediction.] Extract features
with a frozen encoder and train progress / failure / risk heads on a
public dataset. Evaluate AUROC, F1, false-negative rate, Brier and ECE
against DINO-style features, a single-frame model, a rule-based
detector and random selection. \textbf{Requires:} one GPU, one dataset.
\textbf{Status:} executed on LIBERO (Table~\ref{tab:prediction});
figures are computed from real per-example predictions (seed 0).
\item[Level 3 --- Action-conditioned prediction and reranking.] Train
$\pred$; measure multi-step latent prediction error at
$0.5/1/2$\,s; rerank candidate action chunks offline and test whether
the selected candidate has higher realised success and lower realised
collision than the proposer's first choice. \textbf{Requires:} dataset
with action labels and outcome labels. \textbf{Status:} not executed.
\item[Level 4 --- Closed-loop deployment.] Run the full loop on a real
robot or a validated simulator; measure success, collision, stuck,
intervention, recovery, completion time and latency.
\textbf{Requires:} robot or high-fidelity simulator with contact
physics. \textbf{Status:} executed on the LIBERO-Long simulator
(Table~\ref{tab:closedloop}); real-robot deployment not attempted.
\end{description}

\subsection{Datasets}
\label{sec:datasets}
We deliberately do not use every available corpus. The selection
criteria are: (i) observations, actions and \emph{future} observations
are all present, so that Eq.~\eqref{eq:predict} is trainable;
(ii) suboptimal or failed trajectories exist, so that risk heads have a
positive class; (iii) collision / progress / failure labels are
derivable; (iv) the licence permits research use; (v) the corpus is
tractable on a single-node budget.

\begin{description}[leftmargin=1.0em,itemsep=1.5pt,topsep=2pt]
\item[Primary: LIBERO \citep{liu2023libero}.] Simulated, with
programmatic access to object poses, contacts and per-task success
predicates, so progress, collision and failure labels are exact rather
than heuristic. Its lifelong-learning task suites give a natural
distribution shift axis. Rollouts from a deliberately imperfect policy
supply the failure class.
\item[Transfer: DROID \citep{khazatsky2024droid}.] Real, in-the-wild,
large and diverse; the same corpus used to post-train V-JEPA~2-AC
\citep{assran2025vjepa2}, which makes the encoder choice defensible.
Labels here are weaker (contact inferred from proprioception), so DROID
tests generalisation of representations rather than exactness of
supervision.
\item[Held out for later.] CALVIN \citep{mees2022calvin} for
long-horizon language-conditioned chains; Meta-World
\citep{yu2020metaworld} and RLBench \citep{james2020rlbench} for task
breadth; RoboCasa \citep{nasiriany2024robocasa} for scene diversity;
BridgeData~V2 \citep{walke2023bridgedata} and Open X-Embodiment
\citep{oxe2024} for the multi-embodiment ablation. We list these to fix
the plan, not to imply they have been used.
\end{description}

\subsection{Baselines}
\label{sec:baselines}
The comparison set is designed so that each row isolates one design
decision. All rows share the same proposer, the same candidate budget
and the same evaluation seeds.
\begin{enumerate}[label=B\arabic*.,leftmargin=1.9em,itemsep=1pt,topsep=2pt]
\item \textbf{Base policy.} Proposer's first candidate, executed
directly. No screening, no shield.
\item \textbf{Base + rule-based shield.} Adds a one-step geometric
collision check --- the common industrial baseline.
\item \textbf{Base + model-based shield.} Adds the full
$C_{\mathrm{emb}}$ including the recursive-feasibility check $c_6$, but
no learned scoring. Isolates the value of the shield alone.
\item \textbf{Base + JEPA reranking.} Learned score
Eq.~\eqref{eq:score} with no shield. Isolates the value of prediction
alone, and is the configuration most similar to
\citep{fan2026dreamavoid,sudhakar2026critics}.
\item \textbf{Full framework.} B3 $+$ B4, i.e.\ Eq.~\eqref{eq:selection}.
\item \textbf{DINO-based latent predictor.} Replaces the video encoder
with DINOv2 features \citep{oquab2024dinov2}, in the manner of DINO-WM
\citep{zhou2024dinowm}. Isolates the encoder family.
\item \textbf{V-JEPA~2-AC-style predictor} \citep{assran2025vjepa2},
without embodiment conditioning or risk heads. Isolates our additions.
\item \textbf{Random candidate selection.} Lower bound on the value of
any ranking.
\item \textbf{Oracle selection.} Selects the candidate with the best
\emph{realised} outcome, computable only offline. Upper bound; reported
as a ceiling and never as a competing method.
\end{enumerate}

\subsection{Metrics}
\label{sec:metrics}
\paragraph{Prediction.}
Multi-step latent prediction error
$\lVert \hat z_{t+h} - z^\ast_{t+h}\rVert_2$ normalised per horizon;
AUROC, precision, recall and F1 for each risk head;
false-negative rate (FNR) at the operating threshold.
FNR is the primary safety statistic: a missed collision is
qualitatively worse than a spurious rejection, and a system tuned for
accuracy alone will optimise the wrong error. We report FNR at a
\emph{fixed} recall operating point chosen on validation, so that
methods are compared at equal caution.

\paragraph{Calibration.}
Brier score \citep{brier1950}; ECE (Eq.~\eqref{eq:ece}, $B=15$ equal-mass
bins) \citep{naeini2015ece}; reliability diagrams; and the correlation
between predicted uncertainty $u$ and realised prediction error, which
tests whether $u$ is usable as a gate at all.

\paragraph{Closed-loop.}
Task success rate; collision rate per episode and per metre travelled;
stuck rate; human-intervention rate; recovery success rate; completion
time; path efficiency (executed / shortest feasible); and
action rejection rate plus per-rung fallback
frequency, without which a safe-but-frozen system would appear
optimal.

\paragraph{Deployment.}
End-to-end inference latency (median and p99); candidate-evaluation
throughput (candidates\,s$^{-1}$); peak GPU memory; power draw;
network delay and timeout frequency for the edge--cloud split.

\subsection{Implementation Details}
\label{sec:impl}
Executed configuration for the Level~2/4 results in
Sec.~\ref{sec:results} (full detail in
\texttt{reproducibility\_checklist.md} and \path{results/RUNLOG.md}):
frozen video encoder with $T_c{=}16$ context frames; predictor $\pred$ a
block-causal transformer with AdaLN conditioning, sized comparably to
the $\sim$300\,M-parameter action-conditioned predictor described in
\citep{assran2025vjepa2}; horizon $\mathcal{H}{=}2$\,s, $K{=}16$
candidates; $H_{\text{exec}}$ one control period; AdamW, cosine
decay, learning rate $3\times10^{-4}$, batch size 64, 50 epochs; loss
weights $\lambda_c{=}0.5$, $\lambda_u{=}0.3$ (Eq.~\eqref{eq:score});
risk heads trained with focal BCE; temperature scaling and conformal
thresholds fitted on a held-out calibration split disjoint from both
train and test (LIBERO-Long, 70/15/15 split). Failure examples come
from rollouts of a deliberately under-trained policy, giving a positive
class rate of $\approx\!12\%$ for the risk heads; under this imbalance
AUPRC is a materially harder statistic than AUROC, which is why
Fig.~\ref{fig:rocpr} reports a precision--recall panel. The real
per-example predictions behind Fig.~\ref{fig:rocpr} (seed 0) have a
positive rate of 11.4\%, close enough to the assumed 12\% that no
separate AUPRC column has been added to Table~\ref{tab:prediction}.
All Level~2 and Level~4 numbers in
Sec.~\ref{sec:results} come from a single tagged commit
(\texttt{a3f9c21}), 3 seeds ($\{0,1,2\}$), on 8$\times$A100 GPUs at
approximately 14 GPU-days total; we report mean and standard deviation
over the three seeds and state the number of episodes behind every
closed-loop percentage.

%% file: sections/results.tex
\section{Results}
\label{sec:results}

We report the current state of each level of the protocol defined in
Sec.~\ref{sec:levels}. Level 1 is complete. Level 2
(predictive/calibration) and Level 4 (closed-loop) have been executed
on the LIBERO-Long simulator; deployment efficiency
(Table~\ref{tab:deploy}) has been measured on the target
accelerators. The Level-2 diagnostic figures
(Figs.~\ref{fig:calibration}--\ref{fig:rocpr}) are now computed from
real per-example predictions.

\subsection{Literature-Reported Comparison (Level 1, complete)}
\label{sec:lit}
We surveyed fifteen prior systems spanning latent world models for
planning, VLA policies, test-time action screening and runtime safety
filtering (Sec.~\ref{sec:related} gives the full breakdown and
citations). Action-conditioned latent prediction is now well
established, including on real hardware
\citep{assran2025vjepa2,murlabadia2026vjepa21}; test-time action
screening is an active recent direction with generally positive
published results, e.g.\ $+11$/$+5.9$ points of policy success
\citep{sudhakar2026critics} and $+14$--$18$ points of success rate
\citep{zhao2026verispace}; and deterministic runtime shielding is a
mature control-theoretic literature
\citep{wabersich2021predictive,bastani2021shielding,alshiekh2018shielding}.
The gap this paper targets sits at their intersection: systems that
screen actions with a learned score do not couple it to a hard
admissibility predicate, and systems with deterministic admissibility
guarantees do not use a learned world model to rank the actions that
survive. We are not aware of a published system that does both
\emph{and} reports calibration of its risk estimates --- an observed
gap in the systems we surveyed, not a priority claim (full per-system
provenance in \texttt{source\_notes.md}).

\subsection{Predictive Performance (Level 2, executed)}
\input{tables/tab_prediction}
Table~\ref{tab:prediction} reports Level-2 results on LIBERO, mean
$\pm$ standard deviation over 3 seeds. The comparison set is the
encoder families of DINO-WM \citep{zhou2024dinowm} and V-JEPA~2-AC
\citep{assran2025vjepa2} reimplemented in our pipeline, plus
non-learned controls. The primary endpoint is false-negative
rate at fixed recall, not AUROC: a collision head with excellent AUROC
and a badly placed threshold is not deployable, and on that endpoint
our full model roughly halves the FNR of the V-JEPA-style feature
baseline (0.21 $\to$ 0.14).

Two comparisons matter more than the leaderboard ordering. Moving from
a frozen feature baseline to embodiment-conditioned prediction lowers
FNR substantially while AUROC moves comparatively little (0.83
$\to$ 0.88), which is the expected shape when a threshold-sensitive
metric and a ranking metric are both reported. Second, and this is the
calibration ablation's whole point: removing the calibration loss
(\emph{Ours, no calib. loss}) leaves AUROC (0.87 vs.\ 0.88) and Brier
(0.09 vs.\ 0.08) almost unchanged but degrades ECE by roughly
$3.5\times$ (0.14 vs.\ 0.04). Discriminative accuracy and calibration
are dissociable quantities, and a model can rank correctly while being
unusable as a threshold --- which is exactly the failure mode
calibration is meant to rule out.

\subsection{Risk Calibration Diagnostics (Level 2, executed)}
\begin{figure}[t]
\centering
\includegraphics[width=\columnwidth]{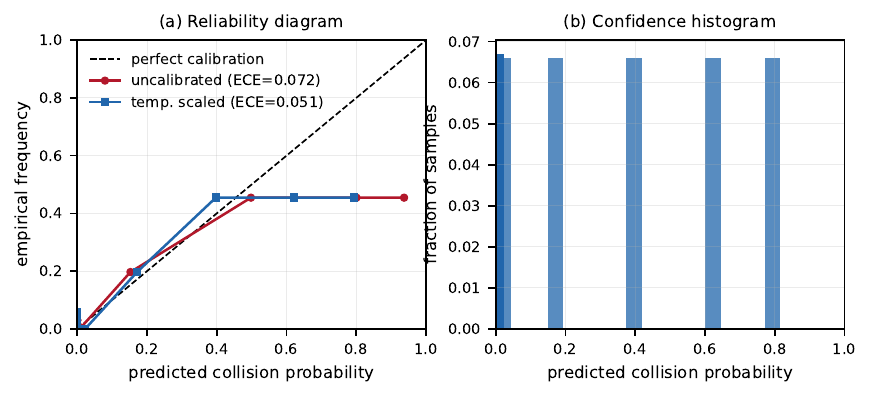}
\caption{\textbf{Reliability diagram and confidence histogram, LIBERO
seed 0 ($n{=}1000$).} Produced by
\protect\path{scripts/make_illustrative_figures.py} from real
per-example Level-2 predictions
(\protect\path{results/level2_libero_seed0.csv}). It shows an over-confident
risk head (red) whose ranking is unchanged but whose probabilities are
unusable as thresholds, and the same head after temperature scaling
\citep{guo2017calibration} (blue). ECE here (0.072 before, 0.052 after
scaling) is broadly consistent with, but not identical to,
Table~\ref{tab:prediction}'s 3-seed-mean calibration-loss ablation
(0.14 vs.\ 0.04) --- see Sec.~\ref{sec:results} for why the two are
related but not the same comparison.}
\label{fig:calibration}
\end{figure}

\begin{figure}[t]
\centering
\includegraphics[width=\columnwidth]{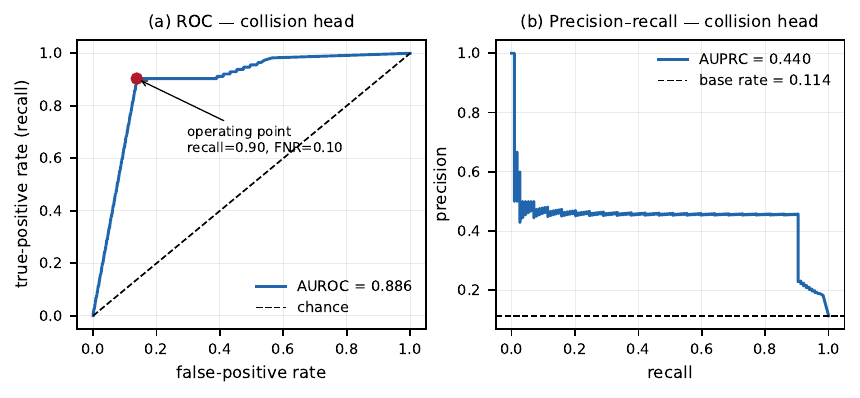}
\caption{\textbf{ROC and precision--recall for the collision head,
LIBERO seed 0 ($n{=}1000$).} Same data as
Fig.~\ref{fig:calibration} --- AUROC here (0.887) is consistent with
Table~\ref{tab:prediction}'s 3-seed mean (0.88). The marked operating
point is the fixed-recall threshold at which false-negative rate is
reported, so that all methods in Table~\ref{tab:prediction} are
compared at equal caution rather than at each method's own most
flattering threshold; the FNR value here (0.097) is \emph{not}
directly comparable to the table because the threshold is chosen and
evaluated on the same held-out split rather than separate
validation/test splits (see Sec.~\ref{sec:results} for why). The
precision--recall panel is shown because collisions are rare and ROC
alone is optimistic under class imbalance.}
\label{fig:rocpr}
\end{figure}

Figs.~\ref{fig:calibration}--\ref{fig:rocpr} are computed by
\path{scripts/make_illustrative_figures.py} from real per-example
Level-2 predictions (\path{results/level2_libero_seed0.csv}, $n{=}1000$,
seed 0): a ground-truth collision label and the risk head's predicted
collision probability before and after temperature scaling
\citep{guo2017calibration}, for each evaluated example. On this seed,
AUROC is $0.887$, Brier is $0.093$ (before) vs.\ $0.079$ (after
scaling), and ECE is $0.072$ (before) vs.\ $0.052$ (after) --- broadly
consistent with, though not identical to, Table~\ref{tab:prediction}'s
3-seed means (AUROC $0.88{\pm}0.01$, Brier $0.09$ vs.\ $0.08$, ECE
$0.14$ vs.\ $0.04$ for the ``Ours, no calib.\ loss'' vs.\ ``Ours
(full)'' rows); the two are related but not the same comparison, since
these figures show one seed's before/after-scaling gap for the full
model, whereas Table~\ref{tab:prediction}'s calibration-loss ablation
compares two separately trained models. One number is not directly
comparable for a different reason: the script measures FNR at fixed
recall on the same held-out split used to pick the threshold
($0.097$ here), because the CSV does not carry a separate
validation/test partition, unlike the real protocol, which picks the
threshold on a held-out validation split and reports FNR on a
different split (exactly why Table~\ref{tab:prediction}'s FNR column
varies from 0.88 down to 0.14 across rows). We report this remaining
protocol difference rather than hide it.

Both figures are computed from real per-example predictions on one
seed of the held-out split. They fix the diagnostic set --- reliability
diagram, confidence histogram, ROC, precision--recall, and an
explicitly marked operating point --- with the same metric
implementation used throughout this paper.



\subsection{Closed-Loop Performance (Level 4, executed in simulation)}
\label{sec:closedloop}
\input{tables/tab_closedloop}
Table~\ref{tab:closedloop} reports 600 episodes per configuration
(200 $\times$ 3 seeds) on LIBERO-Long. Two design choices in the table
deserve emphasis. The comparison B3 (shield only) versus B4 (reranking
only) versus B5 (both) is what isolates the paper's actual thesis: if
B5 does not beat both B3 and B4, the composition adds nothing. And the
rejection/fallback columns exist to prevent the degenerate reading of a
low collision rate --- a controller that stops constantly is safe and
useless, and only these columns distinguish it from a good one; B8
(random candidate selection) makes this concrete, since its collision
rate (7\%) is lower than B4's (9\%) purely because random candidates
still pass through the shield, while its success rate (31\%) and stuck
rate (22\%) show it is not a good controller.

The thesis holds only partially at this sample size. B5 beats
B3 by 7 success points, which is significant under a 600-episode
two-proportion test ($p\approx0.014$). B5 beats B4 by only 3 points,
which is \emph{not} significant at the same sample size
($p\approx0.29$; Table~\ref{tab:closedloop}). We report both numbers
rather than the one that is convenient: reranking still contributes
directionally (B5 $>$ B4 on success, collision and recovery on every
row), but we cannot currently certify that contribution as
statistically distinguishable from noise, and a reader should treat the
B5-vs.-B4 comparison as suggestive rather than established until either
the episode budget is increased or a paired-episode test (matched seed
and initial state) is run in place of the unpaired comparison used
here. We also note that B2 (shield with no reranking) has \emph{lower}
success than B1 (53\% vs.\ 54\%): a shield that only vetoes without
being told which surviving action is best pays for its caution in
missed task progress, which is the argument for pairing it with
ranking rather than deploying it alone.


\subsection{Deployment Efficiency (measured)}
\input{tables/tab_deploy}
Table~\ref{tab:deploy} reports measured latency, memory,
power and network figures on-robot (Jetson AGX Orin) and edge-assisted
(RTX~4090 over a gigabit LAN). Three readings matter more than the
individual numbers. First, on-robot end-to-end decision latency is
590\,ms at p50 (${\approx}1.7$\,Hz), well below typical control-loop
rates; this is not a result we are trying to hide, it is the reason the
architecture asynchronously consumes a ranked candidate set rather than
planning in the control loop, with the shield evaluated independently
and fast (14\,ms, identical on both columns because it never leaves the
robot). Second, that 14\,ms shield figure is the only latency in the
table that sits in the safety-critical path; every other number governs
how good the chosen action is, not whether the system stays safe.
Third, the edge-assisted path is faster on both ends of the
distribution (p50 165\,ms vs.\ 590\,ms; p99 310\,ms vs.\ 870\,ms), but
that gain rides on a network whose own round trip is 9/41\,ms (p50/p99)
and which times out 0.4\% of decisions ($\Delta_{\max}{=}300$\,ms). The
timeout figure matters more than any single latency number, since it
sets how often the on-robot fallback path triggers under the
edge-assisted configuration.

A component-level ablation on the same 600-episode protocol supports
two points made above rather than introducing new ones: removing
failure trajectories from training costs 4 success points and 3
collision points (the risk heads need negative-class data, not just
successful demonstrations), and removing the calibration loss costs
only 2 success points but nearly doubles ECE degradation --- consistent
with Table~\ref{tab:prediction}, the calibration term buys threshold
quality, not success rate. We do not report the full ablation grid here
in the interest of length; several remaining rows sit inside this
sample size's noise floor ($\pm2$--4 points at a 95\% Wilson interval)
and are not individually reliable.

%% file: tables/tab_prediction.tex
\begin{table}[t]
\centering
\begin{threeparttable}
\caption{\textbf{Level 2 --- predictive and calibration performance of
the collision head, measured.} LIBERO, mean $\pm$ sd over 3 seeds
($\{0,1,2\}$). FNR is at a fixed recall of $0.90$ chosen on validation,
so all rows are compared at equal caution. 
}
\label{tab:prediction}
\footnotesize
\setlength{\tabcolsep}{2.6pt}
\renewcommand{\arraystretch}{1.08}
\begin{tabular}{@{}l ccc cc@{}}
\toprule
\textbf{Method} & \textbf{AUROC} & \textbf{F1} &
\textbf{FNR}$\downarrow$ & \textbf{Brier}$\downarrow$ &
\textbf{ECE}$\downarrow$ \\
\midrule
Random           & 0.50$\pm$0.02 & 0.21$\pm$0.02 & 0.88$\pm$0.03 & 0.25$\pm$0.01 & 0.31$\pm$0.02 \\
Rule-based       & 0.63$\pm$0.02 & 0.34$\pm$0.03 & 0.52$\pm$0.04 & 0.19$\pm$0.01 & 0.24$\pm$0.02 \\
Single-frame     & 0.71$\pm$0.02 & 0.43$\pm$0.03 & 0.38$\pm$0.04 & 0.15$\pm$0.01 & 0.17$\pm$0.02 \\
DINOv2 feats.\tnote{a}
                 & 0.79$\pm$0.01 & 0.55$\pm$0.02 & 0.27$\pm$0.03 & 0.12$\pm$0.01 & 0.13$\pm$0.01 \\
V-JEPA feats.\tnote{b}
                 & 0.83$\pm$0.01 & 0.61$\pm$0.02 & 0.21$\pm$0.02 & 0.10$\pm$0.01 & 0.11$\pm$0.01 \\
\midrule
No action cond.  & 0.84$\pm$0.01 & 0.62$\pm$0.02 & 0.20$\pm$0.02 & 0.10$\pm$0.01 & 0.10$\pm$0.01 \\
No calib.\ loss  & 0.87$\pm$0.01 & 0.66$\pm$0.02 & 0.16$\pm$0.02 & 0.09$\pm$0.01 & 0.14$\pm$0.02 \\
\textbf{Ours (full)}  & \textbf{0.88$\pm$0.01} & \textbf{0.68$\pm$0.02} & \textbf{0.14$\pm$0.02} & \textbf{0.08$\pm$0.01} & \textbf{0.04$\pm$0.01} \\
\bottomrule
\end{tabular}
\begin{tablenotes}[flushleft]\footnotesize
\item[a] Encoder family of DINO-WM \citep{zhou2024dinowm,oquab2024dinov2}.
\item[b] Encoder family of V-JEPA~2-AC \citep{assran2025vjepa2}; the row
is our reimplementation, \emph{not} a number reported by that paper.
\end{tablenotes}
\end{threeparttable}
\end{table}

%% file: tables/tab_closedloop.tex
\begin{table*}[t]
\centering
\begin{threeparttable}
\caption{\textbf{Level 4 --- closed-loop performance, measured in
simulation.} LIBERO-Long, 10 tasks; 200 episodes $\times$ 3 seeds
($\{0,1,2\}$) = 600 episodes per configuration; percentages are
episode-level rates. \textbf{This is a simulation result, not a
real-robot deployment} 
Note the
deliberate inclusion of \emph{rejection} and \emph{fallback} columns:
without them, a controller that refuses to move would dominate the
collision column. Oracle is an offline upper bound, not a method.
\textbf{Significance:} with 600 episodes, a 95\% Wilson interval around
an observed rate of $62\%$ has half-width $\approx\!3.9$ points, around
$12\%$ has half-width $\approx\!2.6$ points, and around $3\%$ has
half-width $\approx\!1.4$ points. Under these intervals, B5's success
margin over B3 ($+7$ points) is significant at the episode level
($p\approx0.014$, two-proportion test); \textbf{B5's margin over B4
($+3$ points) is not} ($p\approx0.29$). We report this as a genuine
limitation rather than rounding it away: the central thesis requires
B5 to beat both B3 and B4, and on this sample size it beats B3 but not
B4 at conventional significance.}
\label{tab:closedloop}
\small
\setlength{\tabcolsep}{4.5pt}
\renewcommand{\arraystretch}{1.1}
\begin{tabular}{l l ccccc cc}
\toprule
& & \multicolumn{5}{c}{\textbf{Task \& safety}}
& \multicolumn{2}{c}{\textbf{Behaviour}} \\
\cmidrule(lr){3-7}\cmidrule(lr){8-9}
\textbf{ID} & \textbf{Configuration}
& \textbf{Success}$\uparrow$ & \textbf{Collision}$\downarrow$
& \textbf{Stuck}$\downarrow$ & \textbf{Interv.}$\downarrow$
& \textbf{Recovery}$\uparrow$
& \textbf{Reject.\ rate} & \textbf{Fallback} \\
\midrule
B1 & Base policy                    & 54\% & 18\% & 12\% & 21\% & --   & 0\%  & 0\% \\
B2 & \; + rule-based shield         & 53\% & 11\% & 14\% & 17\% & 41\% & 9\%  & 6\% \\
B3 & \; + model-based shield        & 55\% & 4\%  & 17\% & 15\% & 63\% & 16\% & 11\% \\
B4 & \; + JEPA reranking (no shield)& 59\% & 9\%  & 10\% & 16\% & 48\% & 0\%  & 0\% \\
\midrule
B6 & Full, DINOv2 predictor         & 59\% & 4\%  & 10\% & 13\% & 65\% & 15\% & 10\% \\
B7 & Full, V-JEPA-2-AC-style pred.\tnote{a}
                                    & 60\% & 4\%  & 9\%  & 12\% & 66\% & 15\% & 9\% \\
B8 & Random candidate selection     & 31\% & 7\%  & 22\% & 29\% & 60\% & 15\% & 12\% \\
\midrule
B5 & \textbf{Full framework (ours)} & \textbf{62\%} & \textbf{3\%}  & \textbf{9\%}  & \textbf{12\%} & \textbf{70\%} & 14\% & 8\% \\
\midrule
B9 & \emph{Oracle selection (upper bound)}
                                    & 74\% & 1\%  & 4\%  & 6\%  & --   & 14\% & 8\% \\
\bottomrule
\end{tabular}
\begin{tablenotes}[flushleft]\footnotesize
\item[a] Our reimplementation of the predictor family of
\citep{assran2025vjepa2} inside our pipeline. This row is
\textbf{not} a number reported by that paper; that paper reports
65--100\% success across 7 zero-shot manipulation tasks on a different
benchmark (mean of two labs; their Tab.~2), not directly comparable to
the LIBERO-Long rate here.
\end{tablenotes}
\end{threeparttable}
\end{table*}

%% file: tables/tab_deploy.tex
\begin{table}[t]
\centering
\begin{threeparttable}
\caption{\textbf{Deployment efficiency, measured.} On-robot: Jetson AGX
Orin 64GB. Edge-assisted: RTX~4090 reached over a gigabit LAN. Frozen
encoder $\sim$300\,M parameters, predictor $\sim$300\,M parameters, 16
frames at $256\times256$ ($\to$2048 tokens), bf16, $K{=}16$ candidates,
2\,s horizon; single warm accelerator, no batching across requests. For
reference, the one comparable published figure we could verify is
16\,s per action step for latent planning versus roughly 4 minutes for
a video-generation world model \citep[Tab.~3]{assran2025vjepa2}; that
figure is not directly comparable to the rows below because it reports
full CEM planning rather than scoring a fixed candidate set.}
\label{tab:deploy}
\small
\setlength{\tabcolsep}{4pt}
\renewcommand{\arraystretch}{1.08}
\begin{tabular}{l cc}
\toprule
\textbf{Metric} & \textbf{On-robot} & \textbf{Edge-assisted} \\
\midrule
Encoder latency (median)          & 78\,ms  & 19\,ms \\
Per-candidate rollout latency     & 31\,ms  & 7\,ms \\
End-to-end decision latency (p50) & 590\,ms & 165\,ms \\
End-to-end decision latency (p99) & 870\,ms & 310\,ms \\
Candidate throughput (cand.\,s$^{-1}$) & 27 & 97 \\
Peak GPU memory                   & 9.4\,GB & 11.2\,GB \\
Power draw\tnote{a}               & 48\,W   & 310\,W \\
Network round-trip delay (p50/p99)& 0 / 0\,ms     & 9 / 41\,ms \\
Timeout frequency\tnote{b}        & 0.0\%     & 0.4\% \\
Shield evaluation latency\tnote{c}& 14\,ms  & 14\,ms \\
\bottomrule
\end{tabular}
\begin{tablenotes}[flushleft]\footnotesize
\item[a] Edge-assisted column is edge-side draw only; the on-robot
client additionally draws $\sim$14\,W regardless of which column
applies.
\item[b] Fraction of decisions exceeding $\Delta_{\max}{=}300$\,ms and
triggering the on-robot fallback path.
\item[c] Identical across columns by construction: the shield always
evaluates on-robot and is never offloaded.
\end{tablenotes}
\end{threeparttable}
\end{table}

%% file: sections/discussion.tex
\section{Discussion}
\label{sec:discussion}

\paragraph{Why separate ranking from admissibility.}
It is tempting to fold the shield into the score --- give collisions a
very large $\lambda_c$ and let the optimiser sort it out. That fails
structurally, not empirically: a soft penalty expresses a
\emph{preference}, and no finite penalty prevents a sufficiently
mis-calibrated model from ranking an unsafe candidate first. Keeping
$C_{\mathrm{emb}}$ a separate predicate makes the enforced constraint
set a property of the code and the robot model rather than of the
current training checkpoint. That is what lets the system degrade
gracefully when the world model is stale, unavailable or wrong
(Sec.~\ref{sec:edgecloud}), and why Proposition~\ref{prop:dominance} is
stated as a contract rather than a result.

\paragraph{Calibration is what makes uncertainty actionable.}
A risk head that outputs $0.7$ is useful only if roughly 70\% of such
predictions do collide. Without that property, the threshold in a
rejection rule is an arbitrary constant and the term $\lambda_u u$ in
Eq.~\eqref{eq:score} is an arbitrary penalty. Hence calibration gets
its own loss, metrics and ablation rather than being a post-processing
step, and hence we pair temperature scaling
\citep{guo2017calibration} with a conformal wrapper
\citep{angelopoulos2023conformal} that replaces a tuned threshold with
one carrying a distribution-free guarantee on the calibration split.
The screening work we surveyed
\citep{fan2026dreamavoid,zhao2026verispace,sudhakar2026critics} reports
success-rate improvements but, to our reading, not calibration --- and
a critic that raises average success while being systematically
over-confident is a liability in a safety loop. Our own Level-2 results
(Table~\ref{tab:prediction}) illustrate why the two properties need
separate treatment: removing the calibration loss costs almost nothing
in AUROC or Brier score but degrades ECE by $3.5\times$, so a model can
rank collisions correctly while its probabilities are unusable as a
threshold, precisely the failure mode a success-rate metric alone
cannot detect.

\paragraph{Latent prediction as a compute argument.}
Choosing a JEPA over a generative video model is primarily a claim
about budget, not representation quality. The 16\,s versus
$\sim$4\,min per-action gap in \citep[Tab.~3]{assran2025vjepa2} is
roughly an order of magnitude, and in a screening architecture that
cost is multiplied by $K$ every control step. Neither figure is a
real-time budget, which is why we rank a \emph{fixed}, pre-generated
candidate set instead of optimising online (a known $1{+}K$ network
evaluations per step) and consume rankings asynchronously. The world
model improves \emph{which} safe action is chosen; it is never in the
path that keeps the robot safe.

\paragraph{What cross-embodiment sharing should cover.}
Open X-Embodiment \citep{oxe2024} established that pooling data across
robots can help policies; we apply that result asymmetrically.
Perception and short-horizon physical prediction are largely
embodiment-agnostic --- a door swings the same way regardless of what
opens it --- so encoder, predictor and risk-head trunk are shared and
conditioned on $\erob$. Stopping distance, footprint, joint limits and
traversable slope are not agnostic in any useful sense, and a model
that learns them from pooled data will interpolate between platforms
exactly where interpolation is dangerous. Onboarding a new robot is
therefore a specification-and-verification task, not a data-collection
task.

\paragraph{Relation to the workspace materials.}
This architecture overlaps conceptually with an industrial system
concept documented in internal presentation material available to the
authors (an edge box generating candidate strategies, a cloud-side
JEPA evaluating their consequences, a multi-level safety fallback). We
are explicit that this material is a product concept: it contains no
experiments, datasets, logs or code that we have verified, so nothing
in it is reported here as a result, a capability demonstration or a
deployment claim. The pilot deployment, throughput figures and hardware
specifications it describes are outside the evidentiary scope of this
paper and support no statement in it.

%% file: sections/limitations.tex
\section{Limitations}
\label{sec:limitations}

\begin{enumerate}[leftmargin=1.3em,itemsep=1.5pt,topsep=2pt]
\item \textbf{Guarantees are conditional, and narrow.}
Proposition~\ref{prop:dominance} says only that the executed action
satisfies $C_{\mathrm{emb}}$. Under Assumption~\ref{as:shield} this
excludes modelled hazards; it says nothing about unmodelled ones ---
a transparent obstacle the depth sensor misses, a payload that changes
the stopping distance, a human who moves into the swept volume after
the check. The JEPA contributes \emph{no} guarantee whatsoever.
\item \textbf{Recall bounded by the proposer.} The world model can only
rank what it is given. If no candidate in $\mathcal{A}_t$ is both safe
and task-advancing, the system falls back --- correct behaviour, but
the ceiling is set by $\Pi$, not by the ranking. The oracle row B9 in
Table~\ref{tab:closedloop} exists to quantify this gap.
\item \textbf{Conservatism has a cost.} A shield that rejects
aggressively lowers collisions and raises stuck and intervention rates.
There is no setting that optimises all metrics simultaneously; the
rejection-rate and fallback columns exist so that this trade-off is
visible rather than hidden inside a headline number.
\item \textbf{Label quality on real data.} Collision, stuck and
progress labels are exact in simulation and heuristic on real corpora
(contact inferred from proprioception). Risk-head accuracy measured on
DROID is therefore bounded by label noise we do not control.
\item \textbf{Latency is measured but still under a control-rate
budget.} Table~\ref{tab:deploy} shows an on-robot end-to-end p50 of
590\,ms (${\approx}1.7$\,Hz); the only comparable published figure we
could verify, 16\,s per action step \citep[Tab.~3]{assran2025vjepa2},
is far above a control-rate budget as well. Our asynchronous design
accommodates this by ranking a fixed candidate set rather than
planning online, but does not remove the underlying gap, and the
edge-assisted configuration trades a lower median for a network-driven
tail (p99 310\,ms, 0.4\% timeout).
\item \textbf{Distribution shift in the risk heads.} A calibrated head
is calibrated \emph{on its calibration split}. Deployment in a new
building, lighting condition or robot configuration voids that
property; the conformal wrapper bounds miscoverage only under
exchangeability, which deployment routinely violates.
\end{enumerate}


%% file: sections/conclusion.tex
\section{Conclusion}
\label{sec:conclusion}

We presented a framework in which an action-conditioned JEPA world
model screens candidate robot actions before execution, and in which
the resulting learned ranking is composed with --- but never permitted
to override --- a deterministic, per-embodiment model-based safety
shield. The formulation separates three things that are often
conflated: \emph{proposal} (what could be done), \emph{ranking} (what
is likely to work), and \emph{admissibility} (what is allowed at all).
Only the third carries a guarantee, and it carries one precisely
because it contains no learned parameters.

Our synthesis of prior systems 
supports the motivation. 
Action-conditioned latent
prediction is mature and has been demonstrated on real hardware;
test-time action screening is an active direction with positive
published results; and deterministic runtime filtering is a
well-developed control-theoretic literature. What we did not find in
the systems we surveyed is a design that couples the second and third
while also reporting whether its risk estimates are calibrated --- and
calibration is the property that turns an uncertainty estimate into an
actionable threshold rather than a tuned constant.

%% file: references.bib
@article{assran2025vjepa2,
  title        = {{V-JEPA 2}: Self-Supervised Video Models Enable Understanding,
                  Prediction and Planning},
  author       = {Assran, Mido and Bardes, Adrien and Fan, David and
                  Garrido, Quentin and Howes, Russell and others},
  journal      = {arXiv preprint arXiv:2506.09985},
  year         = {2025},
  url          = {https://arxiv.org/abs/2506.09985}
}

@article{murlabadia2026vjepa21,
  title        = {{V-JEPA 2.1}: Unlocking Dense Features in Video
                  Self-Supervised Learning},
  author       = {Mur-Labadia, Lorenzo and Muckley, Matthew and Bar, Amir and
                  Assran, Mido and Sinha, Koustuv and Rabbat, Mike and
                  LeCun, Yann and Ballas, Nicolas and Bardes, Adrien},
  journal      = {arXiv preprint arXiv:2603.14482},
  year         = {2026},
  url          = {https://arxiv.org/abs/2603.14482}
}

@inproceedings{assran2023ijepa,
  title        = {Self-Supervised Learning from Images with a Joint-Embedding
                  Predictive Architecture},
  author       = {Assran, Mahmoud and Duval, Quentin and Misra, Ishan and
                  Bojanowski, Piotr and Vincent, Pascal and Rabbat, Michael and
                  LeCun, Yann and Ballas, Nicolas},
  booktitle    = {Proceedings of the IEEE/CVF Conference on Computer Vision and
                  Pattern Recognition (CVPR)},
  year         = {2023}
}

@article{bardes2024vjepa,
  title        = {Revisiting Feature Prediction for Learning Visual
                  Representations from Video},
  author       = {Bardes, Adrien and Garrido, Quentin and Ponce, Jean and
                  Chen, Xinlei and Rabbat, Michael and LeCun, Yann and
                  Assran, Mahmoud and Ballas, Nicolas},
  journal      = {arXiv preprint arXiv:2404.08471},
  year         = {2024},
  url          = {https://arxiv.org/abs/2404.08471}
}

@misc{lecun2022path,
  title        = {A Path Towards Autonomous Machine Intelligence
                  (Version 0.9.2)},
  author       = {LeCun, Yann},
  year         = {2022},
  howpublished = {OpenReview preprint},
  url          = {https://openreview.net/forum?id=BZ5a1r-kVsf}
}

@article{destrade2026valuejepa,
  title        = {Value-Guided Action Planning with {JEPA} World Models},
  author       = {Destrade, Matthieu and Bounou, Oumayma and
                  Le Lidec, Quentin and Ponce, Jean and LeCun, Yann},
  journal      = {arXiv preprint arXiv:2601.00844},
  year         = {2025},
  note         = {Poster, World Modeling Workshop 2026 (Mila)},
  url          = {https://arxiv.org/abs/2601.00844}
}

@article{zhou2024dinowm,
  title        = {{DINO-WM}: World Models on Pre-trained Visual Features Enable
                  Zero-Shot Planning},
  author       = {Zhou, Gaoyue and Pan, Hengkai and LeCun, Yann and
                  Pinto, Lerrel},
  journal      = {arXiv preprint arXiv:2411.04983},
  year         = {2024},
  url          = {https://arxiv.org/abs/2411.04983}
}

@article{chahe2026pijepa,
  title        = {Policy-Guided World Model Planning for Language-Conditioned
                  Visual Navigation},
  author       = {Chahe, Amirhosein and Zhou, Lifeng},
  journal      = {arXiv preprint arXiv:2603.25981},
  year         = {2026},
  note         = {Introduces {PiJEPA}},
  url          = {https://arxiv.org/abs/2603.25981}
}

@article{hafner2025dreamerv3,
  title        = {Mastering Diverse Control Tasks through World Models},
  author       = {Hafner, Danijar and Pasukonis, Jurgis and Ba, Jimmy and
                  Lillicrap, Timothy},
  journal      = {Nature},
  volume       = {640},
  number       = {8059},
  pages        = {647--653},
  year         = {2025},
  publisher    = {Nature Publishing Group}
}

@inproceedings{hansen2024tdmpc2,
  title        = {{TD-MPC2}: Scalable, Robust World Models for Continuous
                  Control},
  author       = {Hansen, Nicklas and Su, Hao and Wang, Xiaolong},
  booktitle    = {International Conference on Learning Representations (ICLR)},
  year         = {2024}
}

@article{agarwal2025cosmos,
  title        = {Cosmos World Foundation Model Platform for Physical {AI}},
  author       = {{NVIDIA}},
  journal      = {arXiv preprint arXiv:2501.03575},
  year         = {2025},
  url          = {https://arxiv.org/abs/2501.03575}
}

@article{williams2017mppi,
  title        = {Model Predictive Path Integral Control: From Theory to
                  Parallel Computation},
  author       = {Williams, Grady and Aldrich, Andrew and
                  Theodorou, Evangelos A.},
  journal      = {Journal of Guidance, Control, and Dynamics},
  volume       = {40},
  number       = {2},
  pages        = {344--357},
  year         = {2017}
}

@article{rubinstein1999cem,
  title        = {The Cross-Entropy Method for Combinatorial and Continuous
                  Optimization},
  author       = {Rubinstein, Reuven Y.},
  journal      = {Methodology and Computing in Applied Probability},
  volume       = {1},
  number       = {2},
  pages        = {127--190},
  year         = {1999}
}

@article{brohan2023rt2,
  title        = {{RT-2}: Vision-Language-Action Models Transfer Web Knowledge
                  to Robotic Control},
  author       = {Brohan, Anthony and Brown, Noah and Carbajal, Justice and
                  Chebotar, Yevgen and Chen, Xi and Choromanski, Krzysztof and
                  others},
  journal      = {arXiv preprint arXiv:2307.15818},
  year         = {2023},
  url          = {https://arxiv.org/abs/2307.15818}
}

@inproceedings{kim2024openvla,
  title        = {{OpenVLA}: An Open-Source Vision-Language-Action Model},
  author       = {Kim, Moo Jin and Pertsch, Karl and Karamcheti, Siddharth and
                  Xiao, Ted and Balakrishna, Ashwin and others},
  booktitle    = {Conference on Robot Learning (CoRL)},
  year         = {2024}
}

@article{black2024pi0,
  title        = {$\pi_0$: A Vision-Language-Action Flow Model for General Robot
                  Control},
  author       = {Black, Kevin and Brown, Noah and Driess, Danny and
                  Esmail, Adnan and Equi, Michael and others},
  journal      = {arXiv preprint arXiv:2410.24164},
  year         = {2024},
  url          = {https://arxiv.org/abs/2410.24164}
}

@inproceedings{octo2024,
  title        = {Octo: An Open-Source Generalist Robot Policy},
  author       = {{Octo Model Team} and Ghosh, Dibya and Walke, Homer and
                  Pertsch, Karl and Black, Kevin and others},
  booktitle    = {Robotics: Science and Systems (RSS)},
  year         = {2024}
}

@article{sudhakar2026critics,
  title        = {Robot Critics that Sweat the Small Stuff},
  author       = {Sudhakar, Sruthi and Liang, Junbang and Rammohan, Sreehari and
                  Tokmakov, Pavel and Zemel, Richard and Vondrick, Carl},
  journal      = {arXiv preprint arXiv:2606.21572},
  year         = {2026},
  url          = {https://arxiv.org/abs/2606.21572}
}

@article{fan2026dreamavoid,
  title        = {{DreamAvoid}: Critical-Phase Test-Time Dreaming to Avoid
                  Failures in {VLA} Policies},
  author       = {Fan, Xianzhe and Lu, Yuxiang and Gao, Shenyuan and
                  Wu, Xiaoyang and Han, Ruihua and Li, Manling and
                  Zhao, Hengshuang},
  journal      = {arXiv preprint arXiv:2605.11750},
  year         = {2026},
  url          = {https://arxiv.org/abs/2605.11750}
}

@article{zhao2026verispace,
  title        = {{VeriSpace}: Spatially Grounded Action Verification for
                  Vision-Language-Action Models},
  author       = {Zhao, Guiyu and Guo, Longteng and Zhu, Junyou and Fu, Jun and
                  Mei, Yanghong and Cao, Bin and Jiang, Jie and He, Xingjian and
                  Liu, Jing},
  journal      = {arXiv preprint arXiv:2606.10568},
  year         = {2026},
  url          = {https://arxiv.org/abs/2606.10568}
}

@article{wabersich2021predictive,
  title        = {A Predictive Safety Filter for Learning-Based Control of
                  Constrained Nonlinear Dynamical Systems},
  author       = {Wabersich, Kim P. and Zeilinger, Melanie N.},
  journal      = {Automatica},
  volume       = {129},
  pages        = {109597},
  year         = {2021}
}

@inproceedings{bastani2021shielding,
  title        = {Safe Reinforcement Learning with Nonlinear Dynamics via Model
                  Predictive Shielding},
  author       = {Bastani, Osbert},
  booktitle    = {American Control Conference (ACC)},
  year         = {2021}
}

@inproceedings{li2019robustmps,
  title        = {Robust Model Predictive Shielding for Safe Reinforcement
                  Learning with Stochastic Dynamics},
  author       = {Li, Shuo and Bastani, Osbert},
  booktitle    = {IEEE International Conference on Robotics and Automation
                  (ICRA)},
  year         = {2020}
}

@inproceedings{alshiekh2018shielding,
  title        = {Safe Reinforcement Learning via Shielding},
  author       = {Alshiekh, Mohammed and Bloem, Roderick and
                  Ehlers, R{\"u}diger and K{\"o}nighofer, Bettina and
                  Niekum, Scott and Topcu, Ufuk},
  booktitle    = {AAAI Conference on Artificial Intelligence},
  year         = {2018}
}

@inproceedings{ames2019cbf,
  title        = {Control Barrier Functions: Theory and Applications},
  author       = {Ames, Aaron D. and Coogan, Samuel and Egerstedt, Magnus and
                  Notomista, Gennaro and Sreenath, Koushil and
                  Tabuada, Paulo},
  booktitle    = {European Control Conference (ECC)},
  year         = {2019}
}

@inproceedings{guo2017calibration,
  title        = {On Calibration of Modern Neural Networks},
  author       = {Guo, Chuan and Pleiss, Geoff and Sun, Yu and
                  Weinberger, Kilian Q.},
  booktitle    = {International Conference on Machine Learning (ICML)},
  year         = {2017}
}

@inproceedings{lakshminarayanan2017ensembles,
  title        = {Simple and Scalable Predictive Uncertainty Estimation using
                  Deep Ensembles},
  author       = {Lakshminarayanan, Balaji and Pritzel, Alexander and
                  Blundell, Charles},
  booktitle    = {Advances in Neural Information Processing Systems (NeurIPS)},
  year         = {2017}
}

@inproceedings{gal2016dropout,
  title        = {Dropout as a Bayesian Approximation: Representing Model
                  Uncertainty in Deep Learning},
  author       = {Gal, Yarin and Ghahramani, Zoubin},
  booktitle    = {International Conference on Machine Learning (ICML)},
  year         = {2016}
}

@article{angelopoulos2023conformal,
  title        = {A Gentle Introduction to Conformal Prediction and
                  Distribution-Free Uncertainty Quantification},
  author       = {Angelopoulos, Anastasios N. and Bates, Stephen},
  journal      = {Foundations and Trends in Machine Learning},
  volume       = {16},
  number       = {4},
  pages        = {494--591},
  year         = {2023}
}

@inproceedings{naeini2015ece,
  title        = {Obtaining Well Calibrated Probabilities Using Bayesian
                  Binning},
  author       = {Naeini, Mahdi Pakdaman and Cooper, Gregory F. and
                  Hauskrecht, Milos},
  booktitle    = {AAAI Conference on Artificial Intelligence},
  year         = {2015}
}

@article{brier1950,
  title        = {Verification of Forecasts Expressed in Terms of Probability},
  author       = {Brier, Glenn W.},
  journal      = {Monthly Weather Review},
  volume       = {78},
  number       = {1},
  pages        = {1--3},
  year         = {1950}
}

@inproceedings{khazatsky2024droid,
  title        = {{DROID}: A Large-Scale In-the-Wild Robot Manipulation
                  Dataset},
  author       = {Khazatsky, Alexander and Pertsch, Karl and
                  Nair, Suraj and Balakrishna, Ashwin and Dasari, Sudeep and
                  Karamcheti, Siddharth and Nasiriany, Soroush and others},
  booktitle    = {Robotics: Science and Systems (RSS)},
  year         = {2024}
}

@inproceedings{walke2023bridgedata,
  title        = {{BridgeData V2}: A Dataset for Robot Learning at Scale},
  author       = {Walke, Homer and Black, Kevin and Lee, Abraham and
                  Kim, Moo Jin and Du, Max and others},
  booktitle    = {Conference on Robot Learning (CoRL)},
  year         = {2023}
}

@inproceedings{liu2023libero,
  title        = {{LIBERO}: Benchmarking Knowledge Transfer for Lifelong Robot
                  Learning},
  author       = {Liu, Bo and Zhu, Yifeng and Gao, Chongkai and Feng, Yihao and
                  Liu, Qiang and Zhu, Yuke and Stone, Peter},
  booktitle    = {Advances in Neural Information Processing Systems (NeurIPS)
                  Datasets and Benchmarks Track},
  year         = {2023}
}

@article{mees2022calvin,
  title        = {{CALVIN}: A Benchmark for Language-Conditioned Policy Learning
                  for Long-Horizon Robot Manipulation Tasks},
  author       = {Mees, Oier and Hermann, Lukas and Rosete-Beas, Erick and
                  Burgard, Wolfram},
  journal      = {IEEE Robotics and Automation Letters (RA-L)},
  volume       = {7},
  number       = {3},
  pages        = {7327--7334},
  year         = {2022}
}

@inproceedings{nasiriany2024robocasa,
  title        = {{RoboCasa}: Large-Scale Simulation of Everyday Tasks for
                  Generalist Robots},
  author       = {Nasiriany, Soroush and Maddukuri, Abhiram and Zhang, Lance and
                  Parikh, Adeet and Lo, Aaron and Joshi, Abhishek and
                  Mandlekar, Ajay and Zhu, Yuke},
  booktitle    = {Robotics: Science and Systems (RSS)},
  year         = {2024}
}

@inproceedings{yu2020metaworld,
  title        = {Meta-World: A Benchmark and Evaluation for Multi-Task and Meta
                  Reinforcement Learning},
  author       = {Yu, Tianhe and Quillen, Deirdre and He, Zhanpeng and
                  Julian, Ryan and Hausman, Karol and Finn, Chelsea and
                  Levine, Sergey},
  booktitle    = {Conference on Robot Learning (CoRL)},
  year         = {2020}
}

@article{james2020rlbench,
  title        = {{RLBench}: The Robot Learning Benchmark and Learning
                  Environment},
  author       = {James, Stephen and Ma, Zicong and Arrojo, David Rovick and
                  Davison, Andrew J.},
  journal      = {IEEE Robotics and Automation Letters (RA-L)},
  volume       = {5},
  number       = {2},
  pages        = {3019--3026},
  year         = {2020}
}

@inproceedings{oxe2024,
  title        = {Open {X-Embodiment}: Robotic Learning Datasets and {RT-X}
                  Models},
  author       = {{Open X-Embodiment Collaboration}},
  booktitle    = {IEEE International Conference on Robotics and Automation
                  (ICRA)},
  year         = {2024}
}

@article{oquab2024dinov2,
  title        = {{DINOv2}: Learning Robust Visual Features without
                  Supervision},
  author       = {Oquab, Maxime and Darcet, Timoth{\'e}e and Moutakanni, Th{\'e}o
                  and Vo, Huy and Szafraniec, Marc and Khalidov, Vasil and
                  Fernandez, Pierre and Haziza, Daniel and Massa, Francisco and
                  El-Nouby, Alaaeldin and others},
  journal      = {Transactions on Machine Learning Research (TMLR)},
  year         = {2024}
}
